\documentclass{article}

\PassOptionsToPackage{numbers,sort&compress}{natbib}
\usepackage[preprint]{template/neurips_2026}
\usepackage[utf8]{inputenc} 
\usepackage[T1]{fontenc}    
\usepackage{hyperref}       
\usepackage{url}            
\usepackage{booktabs}       
\usepackage{amsfonts}       
\usepackage{nicefrac}       
\usepackage{microtype}      
\usepackage{xcolor}         
\usepackage{graphicx}
\usepackage{booktabs}
\usepackage{longtable}
 \usepackage{multirow}
 \usepackage{algorithm}
\usepackage{algpseudocode}

\title{NEvo: Neural-Guided Evolutionary Video Synthesis for Dynamic Visual Selectivity}

\author{%
  Yingtian Tang\thanks{Equal contribution.} \\
  EPFL\\
  \texttt{yingtian.tang@epfl.ch} \\
  \And
  Sogand Salehi\footnotemark[1] \\
  EPFL \\
  \texttt{sogand.salehi@epfl.ch} \\
  \And
  Ming Zhou\footnotemark[1] \\
  Johns Hopkins University \\
  \texttt{mzhou52@jhu.edu} \\
  \And
  Amir Zamir \\
  EPFL \\
  \texttt{amir.zamir@epfl.ch} \\
  \And
  Leyla Isik \\
  Johns Hopkins University \\
  \texttt{lisik@jhu.edu} \\
  \And
  Martin Schrimpf \\
  EPFL\\
  \texttt{martin.schrimpf@epfl.ch} \\
}

\begin{document}

\maketitle

\begin{abstract}
    The human brain processes dynamic visual input through hierarchically organized, functionally specialized regions.
    While recent \textit{in silico} brain encoding models can synthesize optimal stimuli to probe selectivity in different brain regions, prior work has been largely limited to static images, leaving dynamic visual processing underexplored.
    We introduce a novel neural-guided video synthesis framework that generates stimuli optimized for target brain regions across visual cortex.
    Our method performs evolutionary search over a structured prompt space, guided by a dynamic encoding model that predicts voxel-level responses to video inputs.
    By maximizing predicted activity for a target ROI, the framework efficiently discovers hyper-activating dynamic stimuli that consistently surpass handcrafted localizer videos.
    The synthesized videos recover known selectivities across ventral, dorsal, and lateral pathways, and further reveal systematic differences in sensitivity to temporal dynamics.
    A searchlight analysis provides new insight into the progression toward increasingly complex social-dynamic features along the lateral stream, further supported by probing with synthesized abstract, non-naturalistic stimuli.
    Taken together, our framework enables \textit{in silico} exploration of dynamic visual selectivity, with new predictions for \textit{in vivo} experiments.
\end{abstract}


\section{Introduction}

Human vision is fundamentally dynamic: the visual system integrates continuously changing input to avoid obstacles, track objects, guide action, and infer others’ intentions as events unfold over time \cite{gibson_ecological_2014,milner2006visual,hasson_hierarchical_2015}.
Dynamic visual processing is supported by hierarchically organized and functionally specialized regions across the visual cortex \cite{dicarlo_how_2012,grill2004human,goodale_separate_1992}. 


Classical accounts emphasize a division between the ventral stream, involved in object and scene recognition, and the dorsal stream, involved in motion processing and visually guided action \cite{mishkin_object_1983, goodale_separate_1992}.
Yet dynamic sensitivity is broadly distributed: MT and V3A encode low- to mid-level motion statistics, the extrastriate body area (EBA) responds to moving bodies and actions, and the superior temporal sulcus (STS) is implicated in biological motion and social interaction, together forming a potential lateral stream \cite{born_structure_2005,jastorff_integration_2012,mcmahon_hierarchical_2023,pitcher_evidence_2021}. Even ventral regions traditionally associated with static category selectivity can be modulated by temporal structure \cite{gilaie2013role,robert_disentangling_2023,tang2025diverse,haxby2020naturalistic}.

Prior research has primarily characterized the visual cortex through region-wise functional selectivity, often using carefully curated stimuli designed to maximally drive specific cortical regions \cite{kanwisher1997fusiform,pitcher_differential_2011}. More recently, deep neural network–based encoding models have become powerful predictors of brain responses to visual input \cite{sartzetaki2025one,schrimpf_integrative_2020,alkhamissi2025language,d2025tribe,gokce2024scaling,tang2025diverse}. 
These models enable large-scale retrieval of stimuli predicted to strongly activate target regions and, increasingly, generative synthesis of stimuli optimized to elicit desired neural responses \cite{cerdas2024brainactiv,hwang2025silico,bashivan_neural_2019,walker_inception_2019,tuckute_driving_2024,luo2023brain}. 
While these approaches have advanced our understanding of object-, face-, body-, and scene-selective responses, they have been limited to the static image domain. 
As a result, they leave open how visual cortex represents the temporal structure of natural input, particularly in regions whose selectivity depends on motion, interaction, or event dynamics.
Addressing this gap requires methods that can systematically generate and evaluate dynamic stimuli across a large joint spatiotemporal configuration space.

\begin{figure*}
    \centering
    \includegraphics[width=\linewidth]{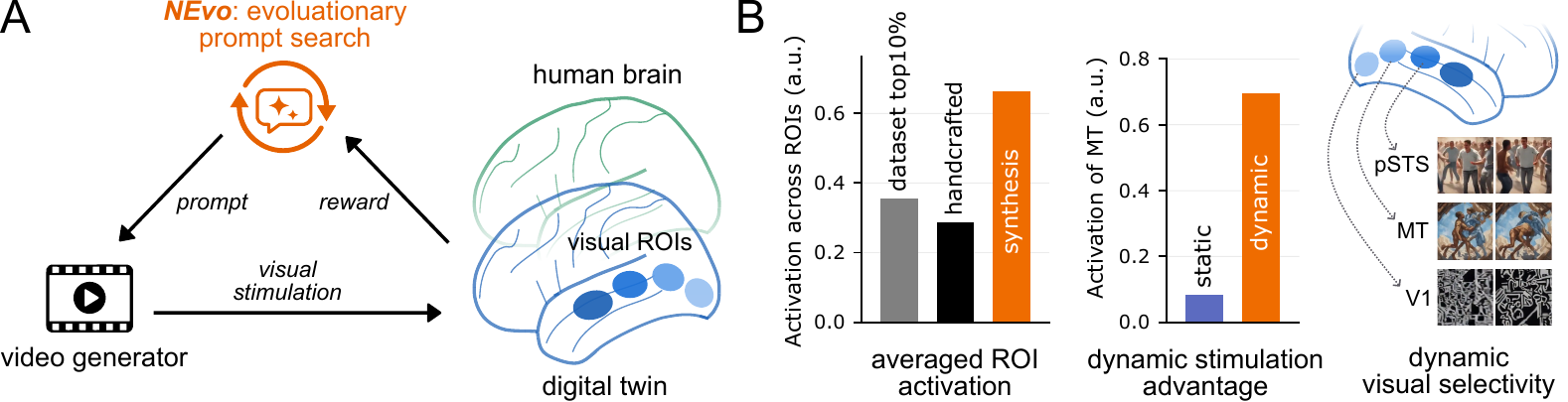}
    \caption{
    \textbf{A. Video synthesis framework.} We propose \emph{NEvo}, a framework that iteratively optimizes video-generation prompts to maximize predicted activation in a target brain region.
    \textbf{B. Overview of results.} By combining dynamic brain encoding with video synthesis, NEvo generates highly activating stimuli, highlights visual dynamics, and enables probing along the lateral visual pathway.
    }
    \label{fig:f1}
\end{figure*}

We introduce \textbf{\emph{NEvo}}, a neural-guided evolutionary video synthesis framework for probing dynamic visual selectivity across the human visual cortex.
As shown in Fig.~\ref{fig:f1}A, NEvo combines a dynamic brain encoding model with an evolutionary search procedure over a structured prompt space.
For a target region of interest (ROI), candidate prompts are rendered into videos by the generation model, scored by the brain encoding model based on predicted ROI activation, and iteratively refined through evolutionary search.
To our knowledge, NEvo establishes the first model-based framework for studying visual selectivity under naturalistic dynamic stimulation.


Our contributions are fourfold (Fig.~\ref{fig:f1}B).
First, we introduce prompt-based neural-guided optimization over interpretable semantic, motion, and event-level descriptors, enabling targeted video synthesis in high-dimensional spatiotemporal space.
Second, we synthesize high-activating stimuli across ventral, dorsal, and lateral visual cortex, exceeding video-dataset retrieval and handcrafted stimuli while recovering known selectivity and revealing regional sensitivity to temporal dynamics.
Third, we use synthesis to characterize dynamic selectivity along the visual hierarchy, revealing a motion-to-social progression along the lateral pathway.
Fourth, we demonstrate controlled synthesis of non-naturalistic stimuli, highlighting NEvo's potential to generate hypothesis-driven stimuli for future experiments.

\section{Related work}

\subsection{Brain-encoding models}
Recent years have seen a surge of work studying the alignment between deep neural networks and the brain \cite{alkhamissi2025language, bao2020map, bakhtiari_functional_2021, caucheteux_brains_2022, choi2023dual, finzi2023single, gokce2024scaling, goldstein_shared_2022, kell_task-optimized_2018, khaligh-razavi_deep_2014, kietzmann_recurrence_2019, nayebi_task-driven_2018, olszewski-kubilius_benchmarking_2019, toneva2019interpreting, d2025tribe, willeke2026omnimouse, wang2025foundation, sartzetaki2025human, sartzetaki2025one, sartzetaki_one_2024, schrimpf_brain-score_2018, schrimpf_integrative_2020, schrimpf_neural_2021, schrimpf_topographic_2024, tang2025diverse, thompson2024zero, wang2019neural, yamins_performance-optimized_2014}.
These studies examine whether, when presented with the same stimuli (e.g., images, videos, audio, or text), the internal activations in the two systems exhibit similar patterns.
From this line of work, brain-encoding models have been developed to accurately predict neural responses.
These models map stimulus representations to brain activity across multiple recording modalities (e.g., EEG \cite{giffordcatalyzing}, fMRI \cite{tang2025diverse, sartzetaki2025one}, electrophysiology \cite{schrimpf_brain-score_2018}) and sensory domains (e.g., vision \cite{schrimpf_integrative_2020}, language \cite{alkhamissi2025language}, audio \cite{li_dissecting_2023}).
For dynamic visual processing, the field has recently benefited from publicly available large-scale video fMRI datasets \cite{mcmahon_hierarchical_2023,lahner_modeling_2024,sava-segal_individual_2023,keles_multimodal_2024,berezutskaya_open_2022}.
These datasets span a broad range of visual dynamics, including extended naturalistic recordings on the scale of tens of minutes \cite{boylecneuromod}.
Models trained on these data can accurately, though imperfectly, predict brain responses to held-out stimuli. \cite{tang2025diverse,sartzetaki_one_2024,d2025tribe}.

\subsection{Model-based characterization of visual selectivity}

The applications of brain-encoding models are only beginning to emerge.
By predicting neural responses to arbitrary stimuli, these models enable systematic probing of functional selectivity. 
In particular, large-scale dataset retrieval or model-guided synthesis of stimuli that maximally activate specific regions \cite{cerdas2024brainactiv, hwang2025silico, bashivan_neural_2019, walker_inception_2019, pierzchlewicz2023energy, luo2023brain, ponce_evolving_2019, ratan2021computational, gu2022neurogen, bao2025mindsimulator, yeung2024neural} offers a far more powerful alternative to traditional small-scale, handcrafted stimulus paradigms \cite{kanwisher1997fusiform, pitcher_differential_2011, epstein1999parahippocampal}.
Dataset retrieval uses the encoding model to screen large collections of visual stimuli and select those that maximally activate a target region of interest (ROI) \cite{hwang2025silico}.
For guided synthesis, a common approach is to use gradients from the encoding model to directly optimize the input stimuli, either under statistical constraints \cite{walker_inception_2019, bashivan_neural_2019} or with the aid of generative priors, such as generative adversarial networks \cite{gu2022neurogen, ratan2021computational} or diffusion models \cite{cerdas2024brainactiv, hwang2025silico, luo2023brain}.
However, to our knowledge, model-based synthesis of dynamic stimuli optimized for target cortical regions has remained unexplored.

\subsection{Dynamic and social visual processing}

Dynamic visual processing is central to natural vision. The dorsal pathway, including MT/MST and V3A, encodes motion, spatial structure, and moving objects or bodies \cite{born_structure_2005,tootell1997functional,simoncelli_model_1998}, while higher parietal regions integrate this information to support attention and visually guided action \cite{orban_parietal_2021,tosoni_sensory-motor_2008,claeys_higher_2003}.
Temporal sensitivity also extends to ventral regions, suggesting that dynamic structure contributes broadly to visual representations \cite{pitcher_differential_2011,robert_disentangling_2023,kucuk_moving_2022}.
The lateral visual pathway, particularly the superior temporal sulcus (STS), links dynamic visual information to social interaction \cite{allison2000social, isik_perceiving_2017, mcmahon_seeing_2023}. Recent work starts to provide evidence for increasing feature complexity along this pathway \cite{pitcher_evidence_2021,mcmahon_hierarchical_2023}, with posterior STS (pSTS) implicated in biological motion, facial movement, and body action \cite{grossman_brain_2002, walbrin_neural_2018, adolphs2003cognitive}, and anterior STS (aSTS) involved in agentive and communicative dynamics \cite{lahnakoski_naturalistic_2012}.


\begin{figure*}
    \centering
    \includegraphics[width=\linewidth]{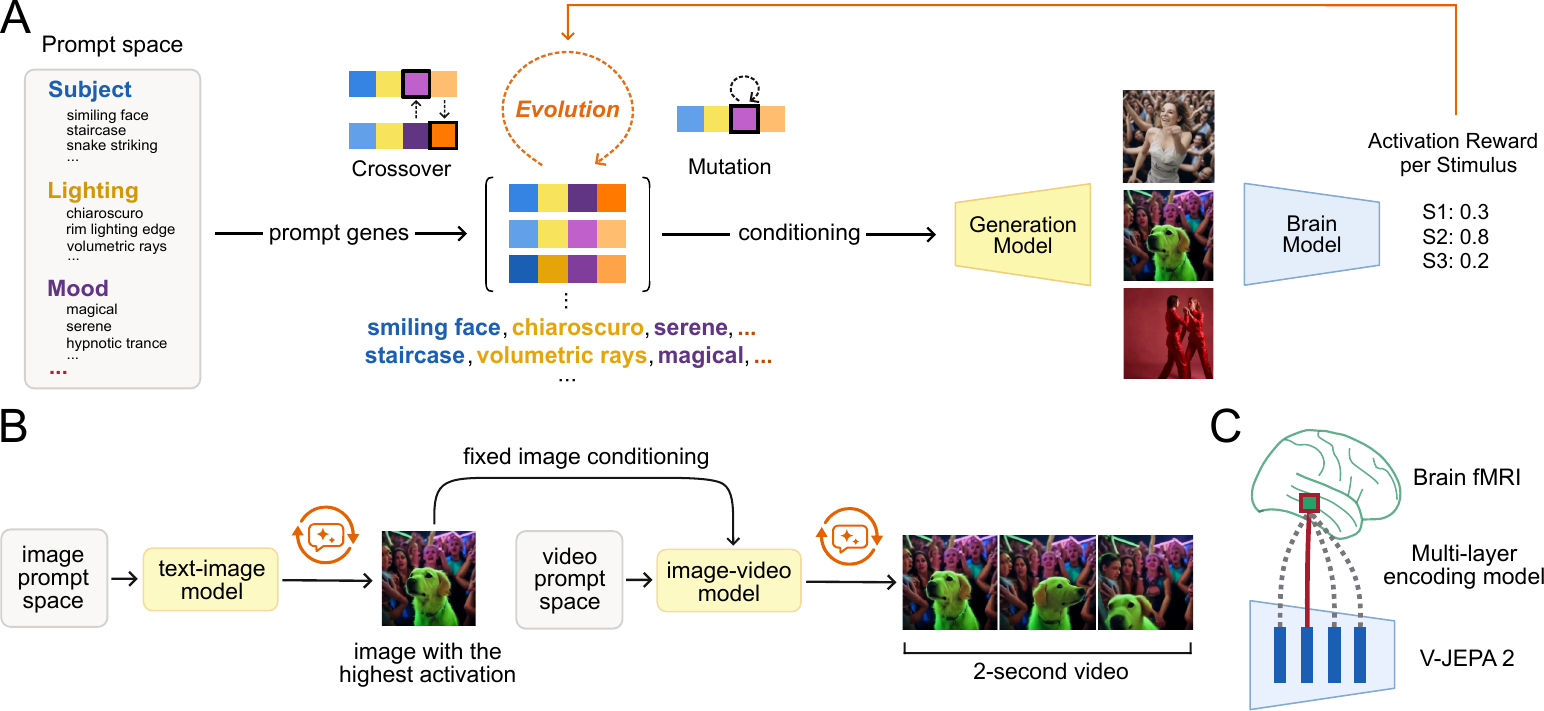}
    \caption{
    \textbf{A. Evolutionary prompting.} Visual attributes are encoded as interpretable genes, decoded into text prompts, and rendered into stimuli by the generation model. Predicted ROI activations from the brain model then guide selection, crossover, and mutation across generations.
    \textbf{B. Two-stage search.} We identify the best-scoring image through image prompt search (Tab.~\ref{tab_image_space_summary}), then use it to initialize image-to-video synthesis with a separate video prompt search (Tab.~\ref{tab_video_space_summary}).
    \textbf{C. Multi-layer brain encoding.} For each voxel, we select the best predictive layer from multiple model layers.
    }
    \label{fig:f2}
\end{figure*}

\section{Method}

\subsection{\emph{NEvo}: neural-guided evolutionary video synthesis}
\label{method:nevo}

We introduce \textit{NEvo}, a neural-guided evolutionary framework for synthesizing dynamic visual stimuli that maximally activate a target region of interest (ROI) \textit{r} under a brain encoding model. 
NEvo searches over a structured prompt space \(\mathcal{P}\), where each \(p \in \mathcal{P}\) describes a candidate video in terms of visual content, scene structure, motion, and temporal dynamics. A video generator \(G\) maps \(p\) to a synthetic video \(x = G(p)\), which is evaluated by a dynamic encoding model \(S_v\) predicting responses for individual voxels \(v\) across visual cortex.

The ROI-level objective is defined as the average predicted response across voxels in the target ROI:
\[
S_r(x) = \frac{1}{|V_r|} \sum_{v \in V_r} S_v(x),
\]
where \(V_r\) denotes the set of voxels in ROI \(r\), and \(S_v(x)\) is the predicted response of voxel \(v\) to video \(x\).
The overall synthesis objective is therefore:
\[
p^\star = \arg\max_{p \in \mathcal{P}} S_r(G(p)),
\]
where \(p^\star\) is the prompt whose generated video \(x^\star = G(p^\star)\) maximally activates ROI \(r\).

As shown in Fig.~\ref{fig:f2}A, the prompt space is structured as a Cartesian product \(\mathcal{P}=\mathcal{A}_1 \times \cdots \times \mathcal{A}_M\), where each \(\mathcal{A}_m\) is a visual attribute category including appearance (subject, mood, lighting, etc) and motion (event structure, camera motion, etc; see Suppl.~\ref{suppl:property-annotation}). A prompt is represented as a gene sequence \(p=(a_1,\ldots,a_M)\), where each \(a_m \in \mathcal{A}_m\) selects one option from attribute \(m\). The selected options are concatenated into a comma-separated text prompt, which is passed to a video diffusion model to generate a two-second video.

We devise an evolutionary search procedure to optimize the synthesis objective (Alg.~\ref{alg:nevo}). At each generation, candidate prompts are rendered into videos and scored using \(S_r\). The top \(P_e = 0.3\) fraction are selected as parents; offspring are produced via  crossover (rate \(P_c = 0.5\)), where random gene-sequence prefixes are swapped between parents, and per-attribute mutation (rate \(P_m = 0.2\)), where attributes are independently resampled. The population size is \(N=20\). This iteratively refines the population toward prompts yielding stronger predicted ROI activation. We compare against random search and hill-climbing baselines (Fig.~\ref{fig:f4}F); full hyperparameter details are in Suppl.~\ref{suppl:hyperparameter}.

The final output of \emph{NEvo} is a ranked set of synthetic videos predicted to strongly activate the target ROI. As the same encoding model and search procedure apply to any ROI or searchlight location, the framework provides a general approach for probing dynamic visual selectivity across cortex.

\begin{algorithm}[t]
\caption{\textsc{NEvo}: Neural-guided evolutionary video synthesis}
\label{alg:nevo}
\begin{algorithmic}[1]
\Require Target ROI \(r\), prompt space \(\mathcal{P}=\mathcal{A}_1 \times \cdots \times \mathcal{A}_M\), video generator \(G\), encoding model \(S_v\), population size \(N\), generations \(T\), elite fraction \(P_e\), crossover rate \(P_c\), mutation rate \(P_m\)
\Ensure Ranked synthetic videos predicted to maximally activate ROI \(r\)

\State Define the ROI score \(S_r(x)=|V_r|^{-1}\sum_{v\in V_r} S_v(x)\), where \(V_r\) is the voxel set of ROI \(r\)
\State Initialize population \(\mathcal{Q}_0=\{p_i\}_{i=1}^{N}\), with \(p_i=(a_{i1},\ldots,a_{iM})\) and \(a_{im}\in\mathcal{A}_m\)

\For{\(t=0,\ldots,T-1\)}
    \State Generate videos \(x_i = G(p_i)\) for all \(p_i \in \mathcal{Q}_t\)
    \State Evaluate candidates using \(s_i = S_r(x_i)\)
    \State Select the top \(P_e\) fraction of candidates as elites
    \State Sample parents from elites and generate offspring via crossover (\(P_c\)) and mutation (\(P_m\))
    \State Form the next population \(\mathcal{Q}_{t+1}\) from elites and offspring
\EndFor

\State \Return Prompt--video pairs ranked by \(S_r(G(p))\)
\end{algorithmic}
\end{algorithm}

\subsection{Two-stage video synthesis}
\label{method:two-stage}

To improve search efficiency, we decompose \(\mathcal{P}\) into static and dynamic subspaces, \(\mathcal{P}=\mathcal{P}_{\mathrm{img}}\times\mathcal{P}_{\mathrm{vid}}\). As shown in Fig.~\ref{fig:f2}B, the first stage searches over \(\mathcal{P}_{\mathrm{img}}\) using a fast image generator \(G_{\mathrm{img}}\), optimizing static content (subject, scene, texture, lighting, etc.). Each candidate \(x_{\mathrm{img}}=G_{\mathrm{img}}(p_{\mathrm{img}})\) is converted into a static video, evaluated using \(S_r\), and the top-scoring candidate is retained as the visual anchor.

The second stage fixes the anchor image and searches over \(\mathcal{P}_{\mathrm{vid}}\), which covers dynamic attributes such as motion profile, temporal rhythm, camera motion, event structure, and interaction type. An image-to-video generator \(G_{\mathrm{vid}}\) maps the anchor and dynamic prompt to a video \(x = G_{\mathrm{vid}}(x_{\mathrm{img}}, p_{\mathrm{vid}})\), evaluated by \(S_r(x)\). This two-stage strategy is efficient because static content and temporal dynamics can be optimized largely independently, avoiding expensive video evaluations for poor content configurations. We use \(T_{\mathrm{img}}=400\) and \(T_{\mathrm{vid}}=200\) evaluations for the image and video stages, respectively.
See Suppl.~\ref{suppl:search-space} for search-space details and Suppl.~\ref{suppl:generation-model} for generator details.

\subsection{Dynamic brain encoding model}

We use V-JEPA 2~\cite{assran2025v} as the feature backbone, motivated by prior large-scale evaluations of brain-aligned video models~\cite{tang2025diverse}. As shown in Fig.~\ref{fig:f2}C, unlike prior work relying on top-level representations such as final CLIP embeddings~\cite{cerdas2024brainactiv, luo2023brain, radford2021learning}, we extract multi-block V-JEPA 2 features, average them over space and time, and fit voxel-wise ridge regressions to fMRI responses (Suppl.~\ref{suppl:brain-model}). Each voxel is assigned the block with the best validation performance, allowing cortical locations to align with different model depths. We train the model-to-brain mapping on two fMRI datasets, \textit{BOLDMoments}~\cite{lahner_modeling_2024} and a social interaction dataset~\cite{mcmahon_hierarchical_2023}, both projected onto the \textit{fsaverage5} cortical surface. Because the model predicts normalized responses, we report activations in arbitrary units (a.u.). For full framework reproducibility, see Suppl.~\ref{suppl:reproduce}.



\begin{figure*}
    \centering
    \includegraphics[width=\linewidth]{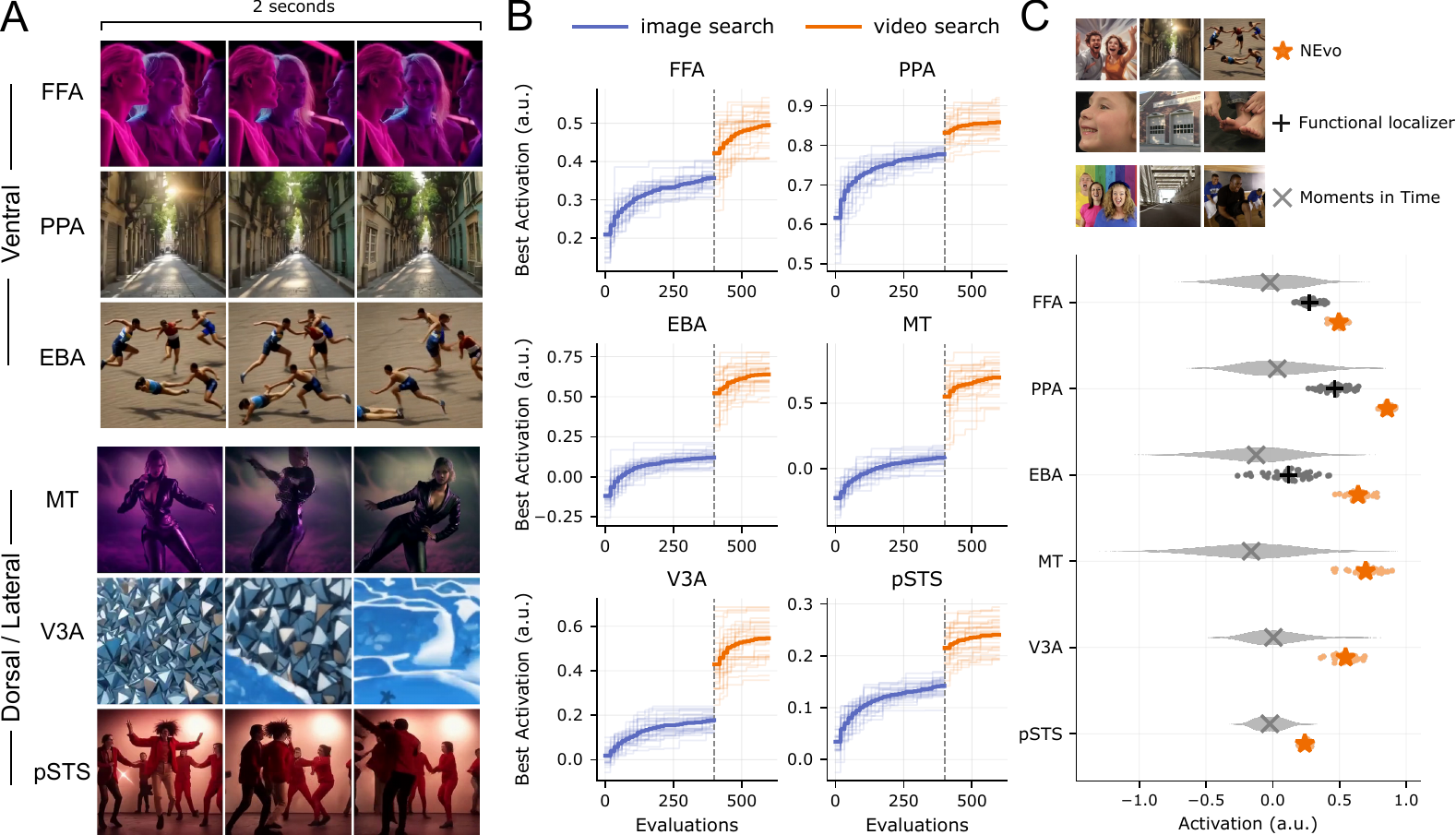}
    \caption{
    \textbf{A. Example synthetic videos from NEvo.} Two-second videos optimized to maximize predicted activation in ventral and dorsal/lateral regions.
    \textbf{B. Two-stage search over iterations.} Optimization trajectories across genetic-search evaluations for each ROI, spanning both image and video search phases. Faded lines denote individual seeds; the solid line denotes the mean.
    \textbf{C. Comparison of ROI activations.} Predicted ROI activations are compared across stimuli from NEvo, dynamic functional localizers \cite{pitcher_differential_2011}, and the Moments in Time dataset \cite{monfort2019moments}. 
    }
    \label{fig:f3}
\end{figure*}

\begin{figure*}
    \centering
    \includegraphics[width=\linewidth]{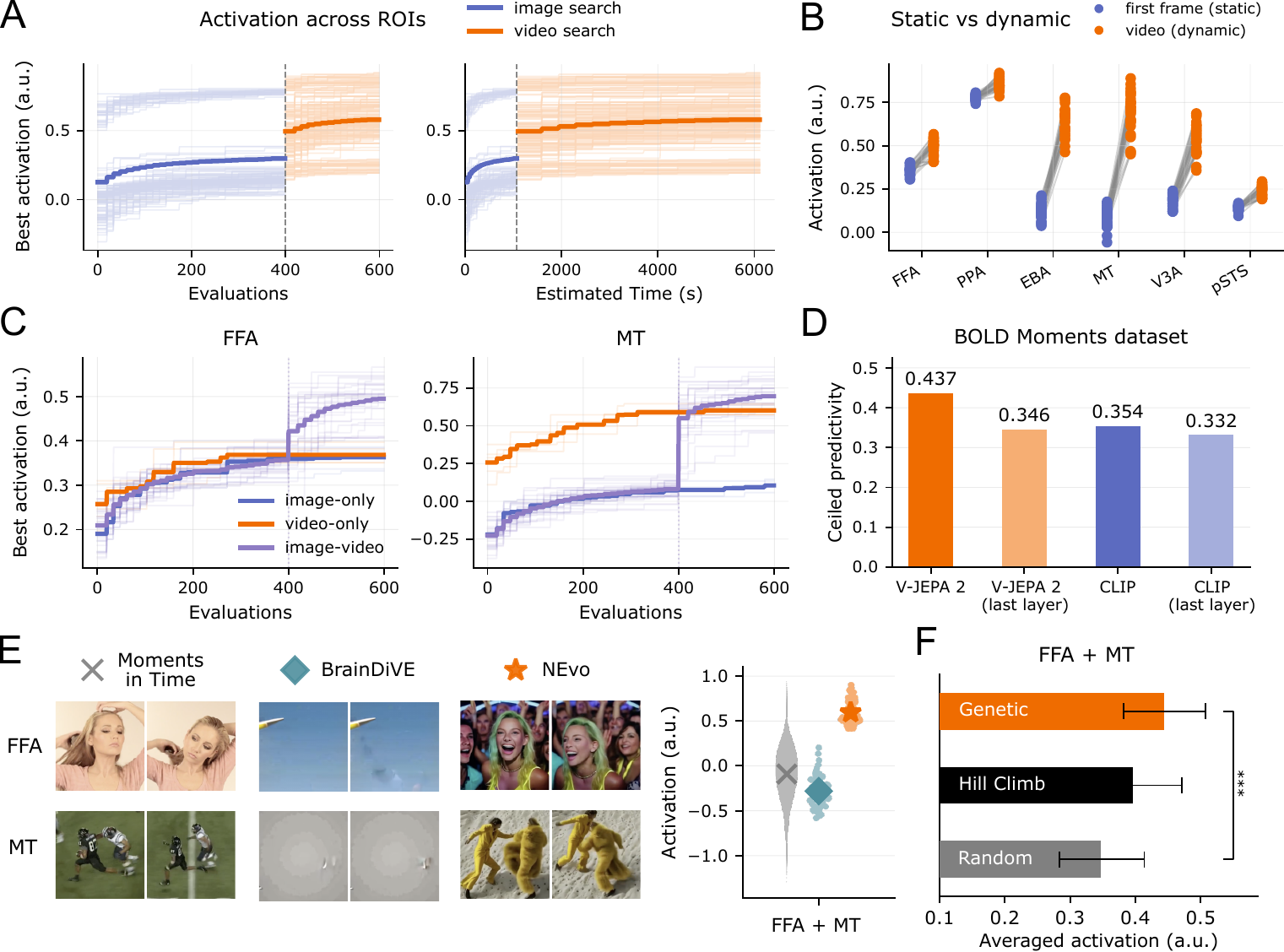}
    \caption{
    \textbf{A. Two-stage search over time.} Optimization trajectories across search evaluations and wall-clock time, aggregated across ROIs.
    \textbf{B. Dynamic stimuli improve activation.} For each ROI, we compare predicted activation between the synthetic video and its first frame presented as a static video.
    \textbf{C. Ablation of two-stage search.} The two-stage strategy outperforms single-stage search baselines in two representative regions.
    \textbf{D. Ablation of encoding models.} Our dynamic multi-layer encoding outperforms the commonly used CLIP last-layer baseline \cite{luo2023brain,cerdas2024brainactiv} and its own last-layer control (Suppl.~\ref{suppl:brain-model}).
    \textbf{E. Ablation of synthesis methods.} NEvo robustly synthesizes high-activating stimuli, whereas the gradient-based BrainDiVE approach \cite{luo2023brain} fails in this dynamic encoding-and-generation setting (additional samples in Fig.~\ref{fig:videos-braindive}).
    \textbf{F. Ablation of search methods.} Genetic search is more robust than hill climbing and random search controls. Error bars show 95\% bootstrap CIs.
    }
    \label{fig:f4}
\end{figure*}

\begin{figure*}
    \centering
    \includegraphics[width=\linewidth]{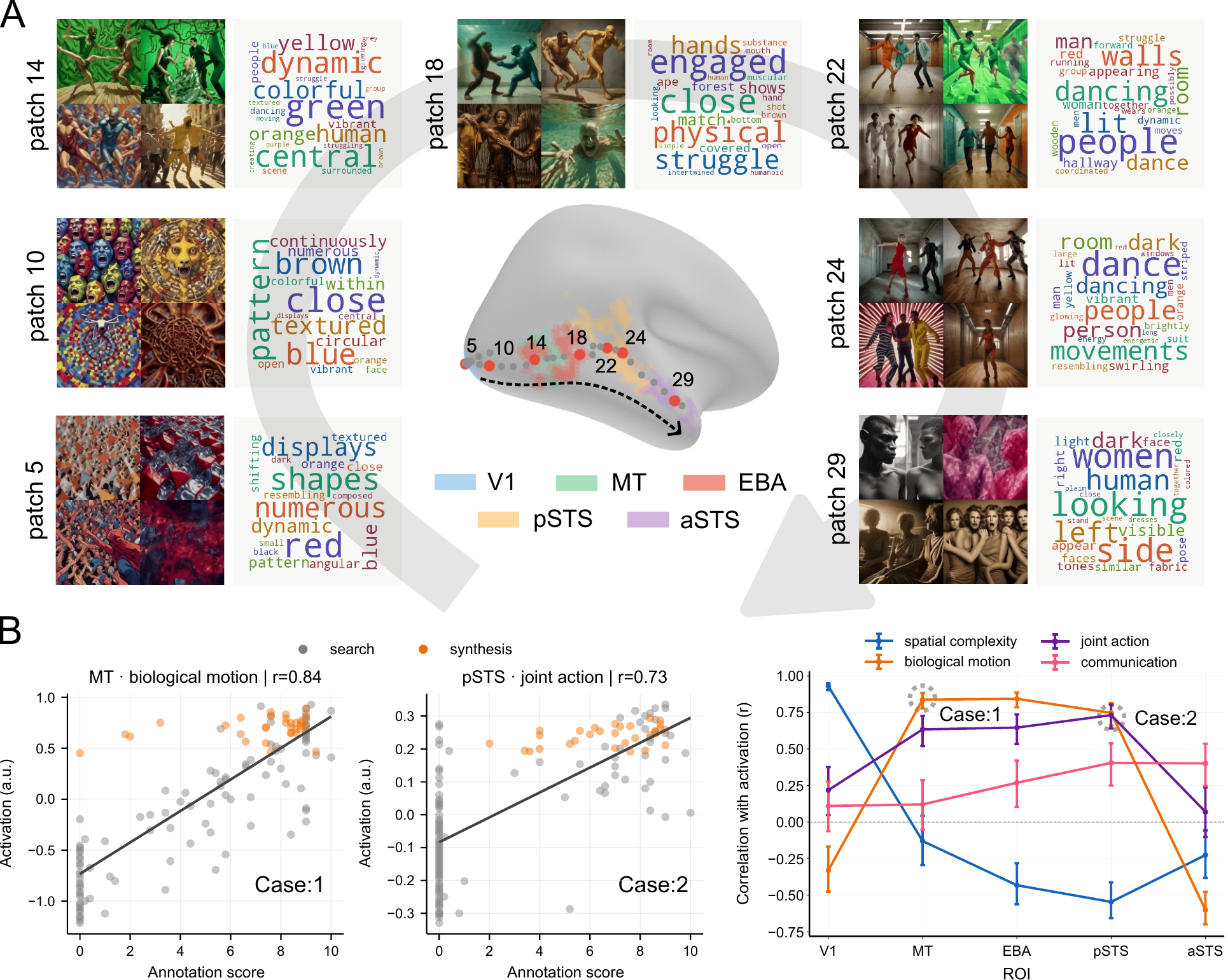}
    \caption{
    \textbf{A. Searchlight synthesis.} Top-synthesis stimuli are shown alongside word clouds from Gemini-annotated descriptions for example searchlight patch along the lateral stream trajectory (V1 to aSTS). Patch indices are marked on the cortical surface.
    \textbf{B. Visual property--activation correlations.}
    Left: Representative correlations between annotated visual-property scores and predicted activations for two example cases: biological motion in MT and joint action in pSTS. Orange points indicate synthesized stimuli and gray points indicate search stimuli from MiT; additional correlations in Fig.~\ref{fig:searchlight-more-correlations}. Right: Correlation profiles across lateral-stream ROIs for visual properties; Error bars denote 95\% CIs derived via 
    Fisher $z$-transformation; full results in Tab.~\ref{tab_property_activation_correlation}.
    }
    \label{fig:f5}
\end{figure*}

\section{Results}

\subsection{NEvo synthesizes hyper-activating videos across visual cortex}

We evaluate the two-stage NEvo on representative ROIs spanning ventral, dorsal, and lateral visual streams, including FFA, PPA, EBA, MT, V3A, and pSTS (ROI definition in Suppl.~\ref{suppl:roi-construction}). 
All candidates are scored by our fixed dynamic encoding model, using the mean predicted response across target-ROI voxels as the objective.
To ensure consistent scoring, all videos are stored as files at a fixed quality setting and reloaded before being passed to the encoding model (Suppl.~\ref{suppl:brain-model}).

Fig.~\ref{fig:f3}A shows representative NEvo videos, which recover expected regional selectivities, including face-like content for FFA, scene-like content for PPA, body-related content for EBA, coherent motion for MT/V3A, and interaction-like dynamics for pSTS (more examples in Suppl.~\ref{suppl:additional-viz}). 
Fig.~\ref{fig:f3}B shows optimization trajectories across random seeds, with each seed yielding one final optimized video. 
Mean predicted ROI activation increases across evaluations in each target region, with substantial gains in both search stages.

In Fig.~\ref{fig:f3}C, we then compare NEvo stimuli with commonly used dynamic functional localizers and videos from Moments in Time (MiT). 
Across ROIs, NEvo reaches high predicted activations relative to both reference sets, achieving the top 99.8\% of the MiT response and 95.8\% of the localizer response on average. 
Together, these results show that \emph{NEvo} efficiently synthesizes high-activating videos that recover known ventral and dorsal/lateral selectivities, while matching or exceeding strong responses from real-video retrieval and dynamic localizers.

\subsection{Dynamic structure enhances regional activation beyond static appearance}

We test whether optimized videos drive ROIs beyond static appearance alone. We track predicted activation across the two-stage search, measured both by search evaluations and wall-clock time, and compare each optimized video with its first-frame static control. Despite its higher computational cost, the video phase further improves activation after image search (Fig.~\ref{fig:f4}A; mean improvement: $0.27 \pm 0.03$, 95\% CI; same CI convention below). 
Across ROIs, optimized videos outperform first-frame controls (Fig.~\ref{fig:f4}B), indicating broad sensitivity to dynamic information, with the strongest effects in motion-related regions (e.g., MT; \(0.61 \pm 0.05\)) and measurable effects in ventral regions (e.g., FFA; \(0.14 \pm 0.02\)).

Finally, the two-stage image-to-video strategy outperforms single-stage search overall (Fig.~\ref{fig:f4}C). 
Relative to direct video synthesis, gains are strong for FFA (\(0.13 \pm 0.04\)) and less significant for MT (\(0.09 \pm 0.14\)), suggesting that FFA benefits from identifying an appropriate appearance anchor before optimizing specific motion patterns, whereas MT is more broadly driven by motion-rich videos.
Together, these results show that temporal structure contributes broadly to visual selectivity and that two-stage synthesis efficiently exploits this dynamic information.

\subsection{Ablating components of the synthesis framework}

We next ablate key components of NEvo to assess design choices (Fig.~\ref{fig:f4}C--F).
In Fig.~\ref{fig:f4}C, we evaluate the two-stage image-to-video search strategy using five seeds for FFA and MT. The two-stage approach consistently outperforms both video-only and image-only search under matched conditions, demonstrating the benefit of decomposing the optimization across spatial and temporal domains.

Fig.~\ref{fig:f4}D compares encoding models by training on a social interaction fMRI dataset \cite{mcmahon_hierarchical_2023} and testing out-of-distribution predictivity on \textit{BOLDMoments} \cite{lahner_modeling_2024} (Suppl.~\ref{suppl:brain-model}). The V-JEPA 2 multi-layer encoding model substantially outperforms CLIP-based baselines, including the multi-layer variant, supporting the need for temporally grounded video representations. Across both models, multi-layer features outperform single-layer counterparts, highlighting the importance of hierarchical encoding.

In Fig.~\ref{fig:f4}E, we compare NEvo with BrainDiVE, a canonical gradient-based synthesis method, for FFA and MT. In the dynamic setting, BrainDiVE fails to generate meaningful stimuli (additional examples in Fig.~\ref{fig:videos-braindive}), with responses falling below the average level of natural videos from MiT (24.0\% of its responses on average).
In contrast, NEvo outperforms BrainDiVE for every available data point, reliably producing high-activating stimuli and indicating greater robustness of evolutionary prompting than gradient-based guidance for video synthesis.


Fig.~\ref{fig:f4}F compares search algorithms using activation across FFA and MT. Genetic search significantly outperforms random search (Cohen’s $d=0.46$; $p<0.001$, paired bootstrap test) and yields higher activation than hill climbing, supporting its effectiveness in high-dimensional prompt space.

\subsection{Model-based synthesis characterizes selectivity across the lateral visual stream}

To use NEvo for dynamic visual selectivity discovery, we apply it along a searchlight trajectory across the lateral visual stream, a recently proposed pathway for social processing that remains only partially characterized \cite{pitcher_evidence_2021,mcmahon_hierarchical_2023}.
We define this trajectory by connecting landmark ROIs from V1 to aSTS along cortical-surface geodesics (shortest paths on the cortical surface; Suppl.~\ref{suppl:roi-construction}), spanning V1, MT, EBA, pSTS, and aSTS. For each searchlight patch along this trajectory, we use NEvo to synthesize high-activating videos.
We then summarize their dominant content using Gemini-2.5-Flash \cite{google_gemini_2026} to generate one-sentence video descriptions and aggregate these descriptions into word clouds.

The synthesized stimuli reveal a systematic progression of feature selectivity along the lateral stream (Fig.~\ref{fig:f5}A). Early visual regions prefer repeated, colorful, high-contrast textures, with descriptions dominated by low-level terms such as ``textured,'' ``numerous,'' and ``red'' (patches 5 and 10). 
Mid-level lateral regions show increasing motion and body content (patch 14), consistent with MT and EBA selectivity. 
Farther along the pathway, stimuli become dominated by physical interactions and synchronized, coordinated body movements, such as ``struggling'' and ``dancing'' (patches 18, 22, and 24). 
In aSTS, synthesized stimuli shift toward face-to-face social contact (patch 29). 
Together, these results show a progressive gradient from low-level spatial structure to body motion and high-level social interaction along the lateral visual stream.

\begin{figure*}
    \centering
    \includegraphics[width=\linewidth]{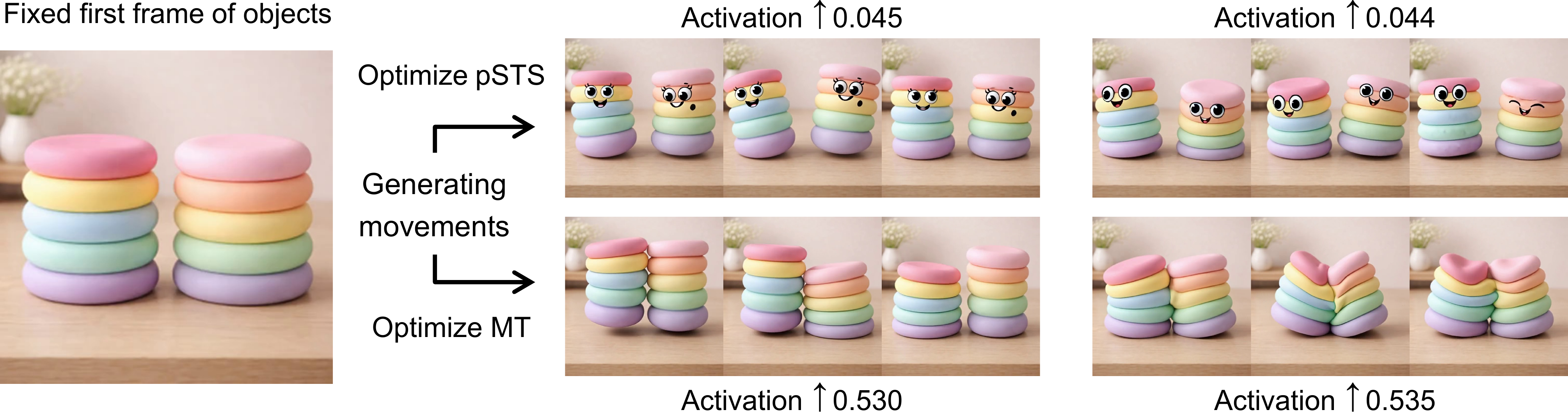}
    \caption{
    \textbf{Controlled synthesis from a non-naturalistic anchor.}
    Starting from a fixed first frame of two stacked plasticine discs, we run video-stage NEvo to maximize predicted pSTS or MT activation. For each ROI, examples show the highest and lowest activation boosts across seeds, measured relative to the static first-frame baseline. pSTS optimization introduces face-like features and coordinated interactions, whereas MT optimization yields motion-rich dynamics with less biological or social content, consistent with Fig.~\ref{fig:f5}. 
    Additional examples and controls are shown in Fig.~\ref{fig:s12}.
    }
    \label{fig:f6}
\end{figure*}

\subsection{Synthesis stimuli induce high activation through region-specific visual properties}

We further explicitly characterize the functional relevance of the synthesized stimuli. We again use Gemini-2.5-Flash to annotate low-level structure, motion, body, scene, and social-interaction properties for both synthesized videos and naturalistic videos sampled uniformly across each ROI’s predicted activation range (Suppl.~\ref{suppl:property-annotation}). 

We compute Pearson correlations between each property and model-predicted activation, and overlay synthesized stimuli on the naturalistic video distributions (Fig.~\ref{fig:f5}B left).
Representative ROIs show strong property–activation correlations: MT with biological motion ($r=0.84$, $p<0.001$), pSTS with joint action($r=0.73$, $p<0.001$).
Across ROIs, synthesized stimuli consistently occupy the high-property, high-activation extreme, suggesting that \emph{NEvo} converges on functionally characteristic stimuli rather than incidental outliers.
Along the lateral pathway, property--activation correlations reveal a gradual transition toward increasingly complex social features, peaking in pSTS, shifting toward less motion-focused communicative features in aSTS (Fig.~\ref{fig:f5}B, right).

As a controlled application of \emph{NEvo}, we fix the static appearance to a first frame containing two abstract objects and optimize only the video dynamics for different ROIs across multiple random seeds (Suppl.~\ref{suppl:controlled_synthesis}). As shown in Fig.~\ref{fig:f6}, pSTS optimization introduces face-like features and coordinated interactions, consistent with social-dynamic selectivity, whereas MT optimization yields translation- and oscillation-rich motion with fewer biological or social attributes. These results show that NEvo can dissociate region-specific dynamic selectivity from abstract anchors, providing a controlled paradigm for generating targeted, hypothesis-driven stimuli.



\section{Discussion}

We introduced \emph{NEvo}, a neural-guided evolutionary framework for synthesizing dynamic stimuli optimized for target regions across human visual cortex.
By coupling a dynamic brain-encoding model with structured prompt-space search, \emph{NEvo} addresses the substantially larger temporal search space of video synthesis, moving model-based stimulus design beyond static images.
Across visual cortex, synthesized videos recover expected selectivities and elicit strong predicted responses, exceeding established localizers even in well-characterized regions such as FFA.
Furthermore, \emph{NEvo} generates testable hypotheses for less-characterized regions, revealing structure along the lateral visual stream.

\emph{NEvo} formulates video synthesis as black-box optimization over interpretable semantic, motion, and event-level descriptors. This formulation is critical for dynamic vision, where the search space is substantially larger and more complex than in static images. Compared with direct gradient-based optimization, evolutionary prompting constrains synthesis toward semantically meaningful and temporally coherent videos, while retaining a flexible search procedure that can target different cortical regions.


Our results show that temporal structure broadly shapes visual selectivity, with optimized videos outperforming first-frame controls in motion-sensitive regions such as MT, and even in ventral regions such as FFA. 
The two-stage image-to-video strategy is especially effective in this setting: it first identifies strong appearance anchors, such as faces for FFA, and then optimizes anchor-specific motion dynamics, improving efficiency while avoiding many uninformative video generations.

Beyond predefined ROIs, \emph{NEvo} enables model-based discovery of dynamic selectivity across the cortical surface.
Along the lateral visual stream, searchlight-level synthesis and activation-property analysis reveal a progression toward increasingly complex socially relevant features, smoothly transitioning from 
motion-rich, body-involved physical movement to highly 
synchronized, coordinated joint interaction, and finally to face-centered, motion-suppressed communicative content.
Controlled synthesis further dissociates these region-specific preferences, even using abstract objects.
Together, these findings provide convergent evidence for functional gradients while revealing fine-grained structure beyond prior ROI-based observations \cite{pitcher_evidence_2021,mcmahon_hierarchical_2023, deen_functional_2015}.

\paragraph{Limitations.}
Several limitations remain. 
\emph{NEvo} depends on the accuracy and coverage of the underlying encoding model, so synthesized stimuli may reflect both neural selectivity and model bias; future validation with human neuroimaging experiments is therefore necessary.
The current model is vision-only, limiting characterization of regions such as aSTS, where communication often relies on speech and audiovisual cues.
Finally, the structured prompt space improves interpretability but may omit relevant visual dimensions not captured by the schema.

\paragraph{Broader impact.}
\emph{NEvo} may benefit neuroscience and clinical research by enabling scalable, controlled design of dynamic stimuli and more reproducible experiments. Potential risks include amplifying biases from framework components, including video generators and encoding models.

Future work should incorporate multimodal encoding models, richer video generators, and search spaces suited to longer event structures. 
Critically, closed-loop neuroimaging should test whether \emph{NEvo}-synthesized videos elicit the predicted human brain responses, potentially extending neural-guided synthesis from mapping healthy brain function to designing targeted clinical interventions.

\newpage

\bibliographystyle{plainnat}
\bibliography{references,references_add}

\section{Acknowledgement}
This work was supported by the Swiss National Science Foundation (SNSF) under Grant No. 10.003.772. This work was also supported under project ID 43 as part of the Swiss AI Initiative, through a grant from the ETH Domain, with computational resources provided by the Swiss National Supercomputing Centre (CSCS) on the Alps infrastructure. This work has also received funding from the Swiss State Secretariat for Education, Research and Innovation (SERI).

\appendix

\setcounter{figure}{0}
\setcounter{table}{0}

\renewcommand{\thefigure}{S\arabic{figure}}
\renewcommand{\thetable}{S\arabic{table}}

\renewcommand{\theHfigure}{suppfig.\arabic{figure}}
\renewcommand{\theHtable}{supptable.\arabic{table}}

\section{Additional visualizations}
\label{suppl:additional-viz}

\begin{figure*} 
    \centering 
    \includegraphics[width=\linewidth]{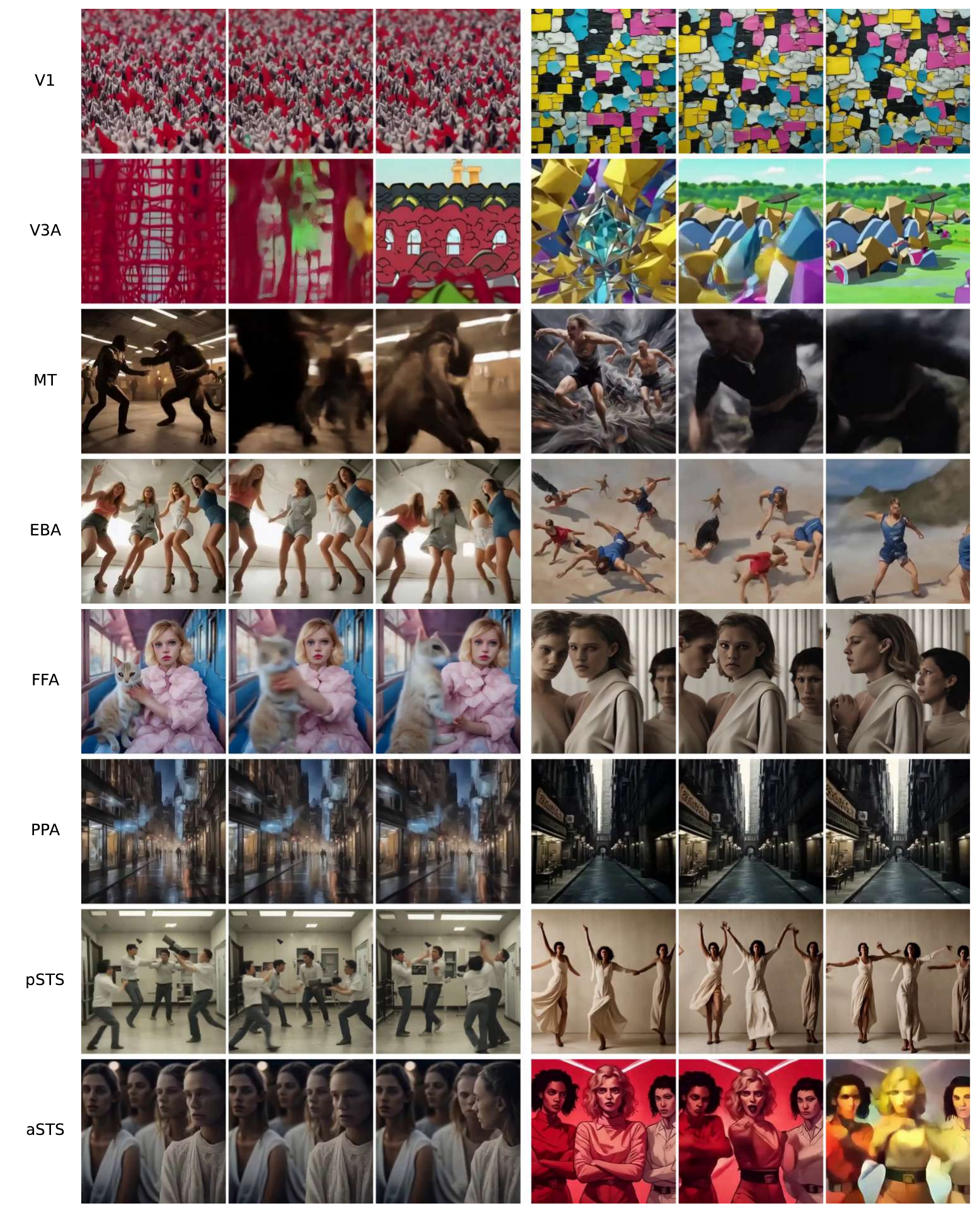} \caption{Additional examples of synthetic videos from NEvo. Two-second videos optimized  to maximize predicted activation in different ROIs.} 
    \label{fig:videos-additional} 
\end{figure*}

\begin{figure*} 
    \centering 
    \includegraphics[width=\linewidth]{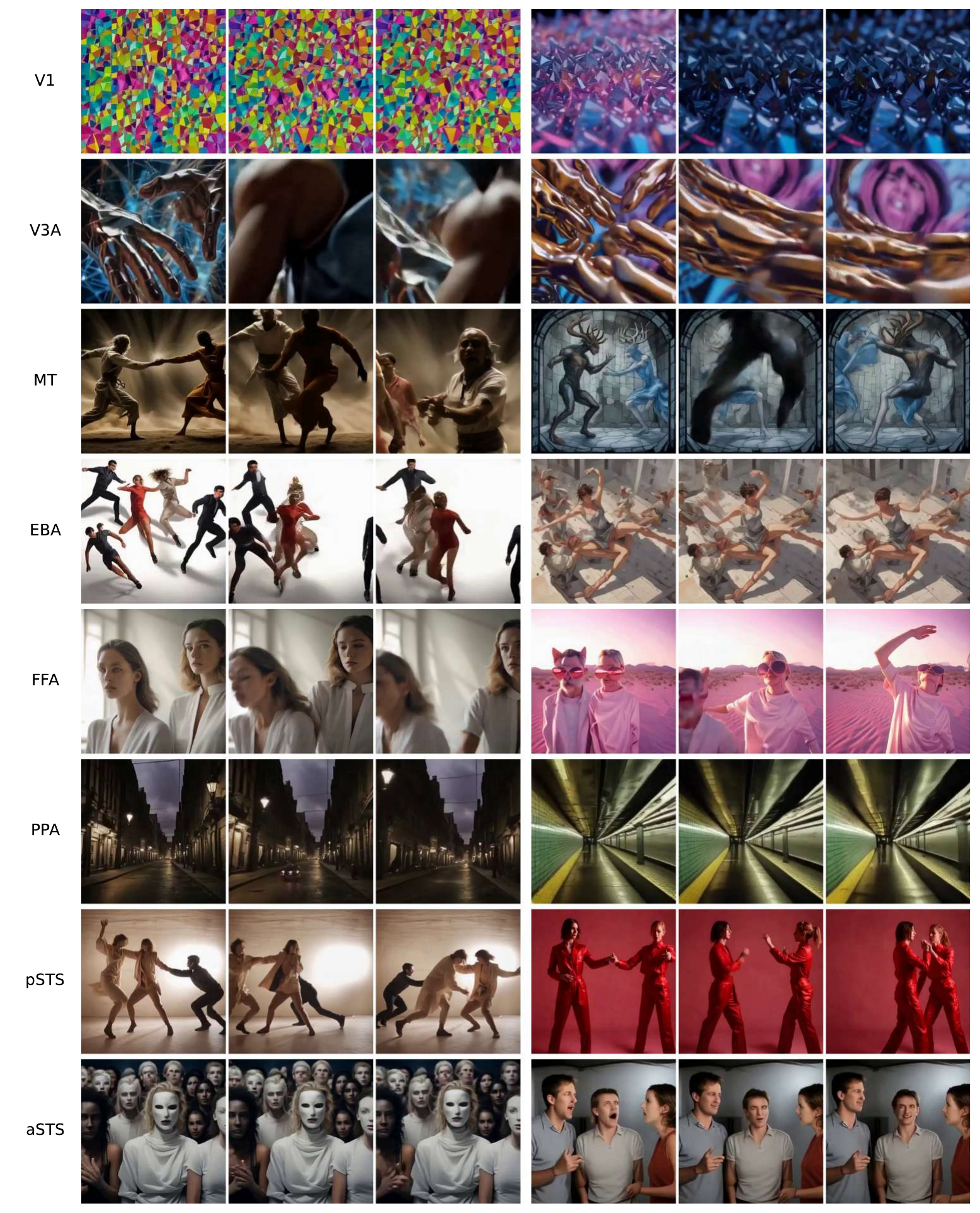} \caption{Additional examples of synthetic videos generated by NEvo, continuing Fig.~\ref{fig:videos-additional}.}
    \label{fig:videos-additional2} 
\end{figure*}

\begin{figure*} 
    \centering 
    \includegraphics[width=\linewidth]{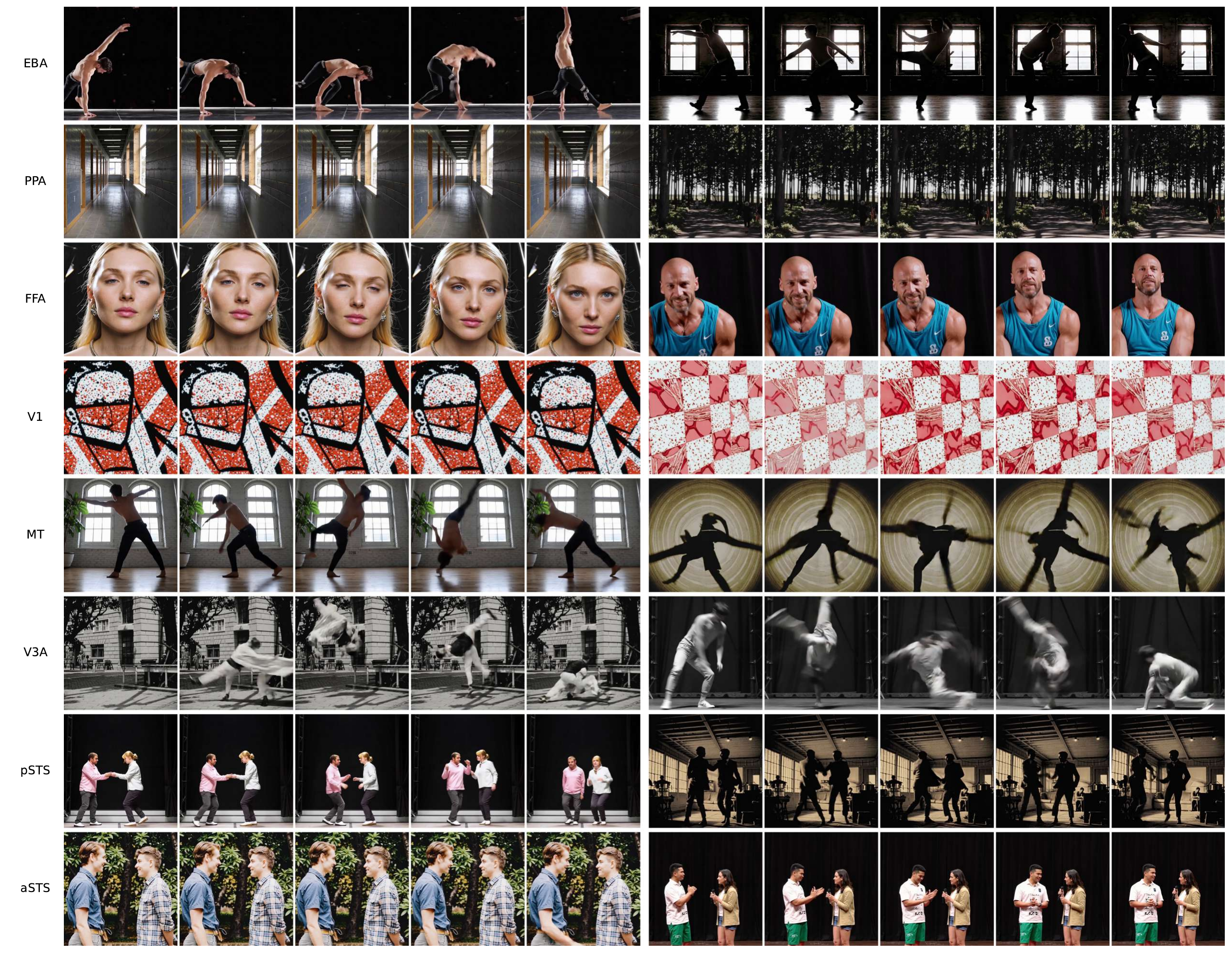} 
    \caption{Additional examples of synthetic videos generated by NEvo with prompts reinforced for high quality. For better presentation, we show five sampled frames from each video.} 
    \label{fig:videos-high-quality} 
\end{figure*}

We provide additional qualitative examples of synthetic videos generated by \emph{NEvo}. Figs.~\ref{fig:videos-additional} and \ref{fig:videos-additional2} show additional two-second videos optimized for different target ROIs.
Fig.~\ref{fig:videos-high-quality} shows examples generated with improved prompt-quality constraints and higher-resolution rendering.

\begin{figure*}
    \centering
    \includegraphics[width=\linewidth]{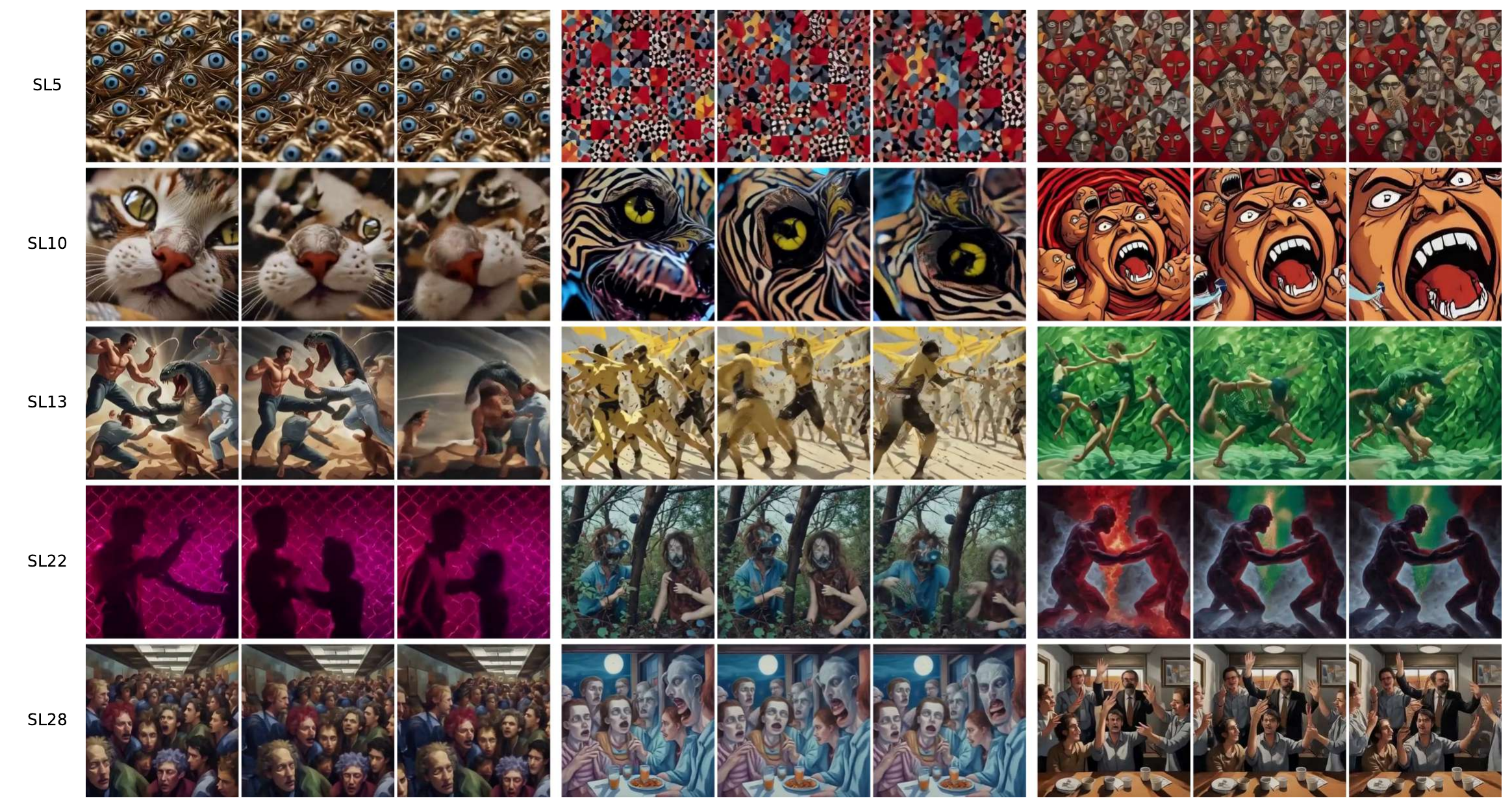}
    \caption{Synthetic stimuli from \emph{NEvo} along the lateral visual stream, highlighting combinations of visual features that are unlikely to appear in natural video datasets. Each row corresponds to a searchlight location (see Fig.~\ref{fig:f5}A), with three independent seeds shown per location. SL5 (SL stands for search light), near V1, produces dense repeating textures and pattern fields with high spatial frequency content. SL10, situated between V1 and EBA, generates high-contrast animal textures and close-up predator faces, reflecting a transition from low-level edge and texture processing toward body and object selectivity. SL13, near EBA and MT, consistently produces scenes of multiple figures engaged in vigorous physical interaction such as fighting and dancing with circular movements. SL22, overlapping pSTS, favors silhouetted dyadic interactions in dramatic lighting, emphasizing the social configuration over identity or texture. SL28, near aSTS, generates crowded social scenes with multiple expressive faces and group interactions. Across locations, the synthesized stimuli reflect known functional preferences of each region while combining features, such as abstract social configurations without identifiable faces, or pattern fields without natural objects, that are rare or absent in naturalistic video datasets.}
    \label{fig:videos-lateral-stream}
\end{figure*}

Fig.~\ref{fig:videos-lateral-stream} provides additional stimuli from the lateral-stream searchlight analysis. These examples illustrate how NEvo synthesizes different feature combinations along the cortical hierarchy, ranging from high-frequency textures near early visual cortex to body motion, dyadic interaction, and crowded social scenes in higher lateral temporal regions. Notably, several synthesized stimuli combine visual features that are uncommon in natural video datasets, such as abstract social configurations without clear identity cues or dense pattern fields without natural objects.

\begin{figure*} 
    \centering 
    \includegraphics[width=\linewidth]{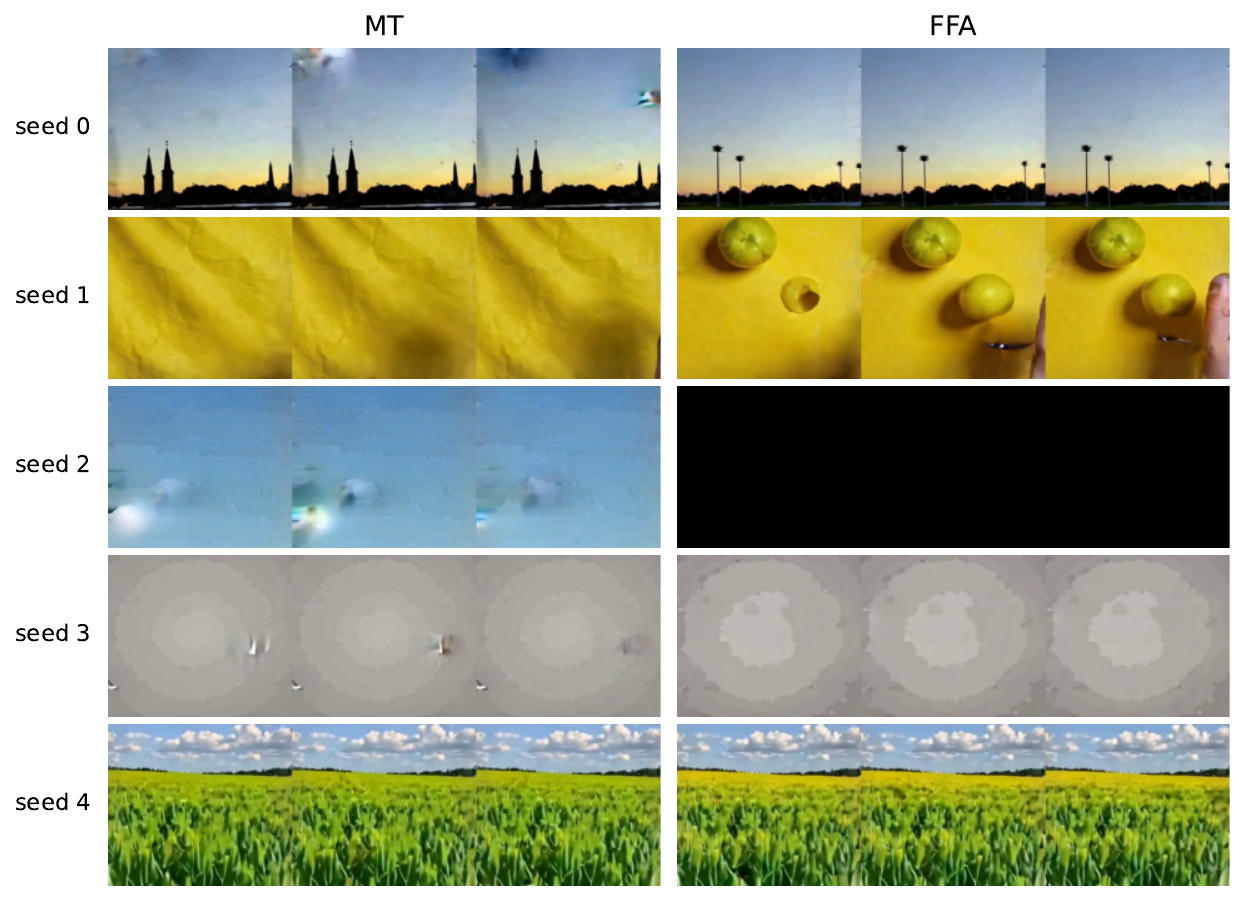} \caption{Additional examples of synthetic videos from BrainDiVE, using the same models specified in Suppl.~\ref{suppl:model-choices}.} 
    \label{fig:videos-braindive} 
\end{figure*}

In Fig.~\ref{fig:videos-braindive}, we show additional samples from applying BrainDiVE to dynamic stimulus synthesis. Many samples contain artifacts introduced by gradient-based modification, and none reflects the target ROI selectivity.

\section{Concrete model choices in \emph{NEvo}}
\label{suppl:model-choices}

\subsection{Generation models}
\label{suppl:generation-model}

For the image search stage, we use SDXL-Turbo~\cite{sdxlturbo}, a distilled single-step variant of Stable Diffusion XL. Images are generated at $512 \times 512$ resolution in a single inference step with guidance scale 0.0.

For the video search stage, we use LTX-Video 2.3~\cite{ltxvid,ltx23}, a 22B distilled diffusion transformer with a spatial upsampler and Gemma-3 12B~\cite{gemma3} as the text encoder. Videos are generated at $256 \times 256$ resolution, 24 fps, and 2 seconds duration. For image-to-video generation, the anchor image from the image search stage is provided as a conditioning input at frame index 0 with strength 1.0. For text-to-video generation, no image conditioning is used. All videos are generated with fp8 quantization.

\begin{figure*} 
    \centering 
    \includegraphics[width=.7\linewidth]{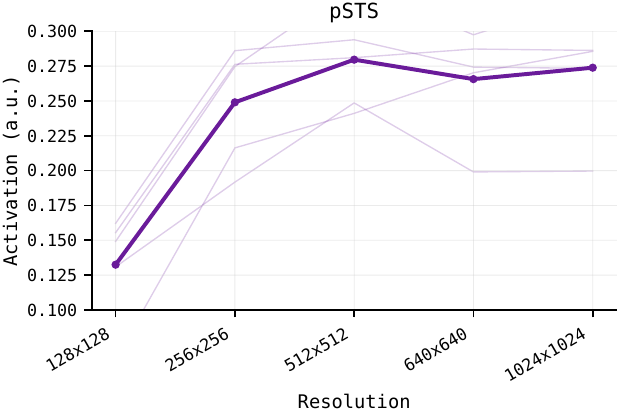} \caption{Best predicted activations obtained by running NEvo at different video generation resolutions, averaged over 5 seeds. Our default setting, described in Suppl.~\ref{suppl:hyperparameter}, uses a resolution of $256 \times 256$.} 
    \label{fig:videos-resolution} 
\end{figure*}

In general, higher generation resolutions yield higher predicted brain responses (Fig.~\ref{fig:videos-resolution}), even though the brain encoding model normalizes all inputs to the same quality (Suppl.~\ref{suppl:brain-model}). The settings above were chosen to balance performance with computational cost.

\subsection{Brain-encoding models}
\label{suppl:brain-model}

To build the V-JEPA 2 encoding model, we use all videos and corresponding fMRI response betas from \textit{BOLDMoments} \cite{lahner_modeling_2024} and a social interaction dataset \cite{mcmahon_hierarchical_2023}. Each video is associated with a single beta volume; therefore, we predict one response vector per video rather than a time-resolved response.
We project the volumetric data onto the \textit{fsaverage} surface and normalize responses per voxel within each dataset.

Videos are input to V-JEPA 2 at 12 fps. We extract features from \texttt{backbone.blocks.\textit{i}.attn}, with \(i \in \{13,15,17,19,21,23\}\). Each feature tensor has dimensions time \(\times\) height \(\times\) width \(\times\) channels. For each layer, we average features over time, height, and width, and independently regress the resulting feature vector onto each voxel using ridge regression. The ridge penalty is selected by cross-validation over 17 logarithmically spaced values between \(10^{-8}\) and \(10^{8}\). We split the data into 90\% training and 10\% validation sets, and assign each voxel to the layer with the best validation performance.

In Fig.~\ref{fig:f4}D, the CLIP-based model is constructed by extracting features from individual video frames using the last six layers of the CLIP vision encoder, concatenating them into a single feature representation, and applying the same averaging and regression procedure. For the ``last-layer'' variants, we use only the final layer rather than the multi-layer feature set described above. For this particular figure, to assess out-of-distribution generalization in this analysis, models are trained on the social interaction dataset and tested on \textit{BOLDMoments}.
The final ceiled predictivity is measured as the Pearson correlation between predicted and measured responses, normalized by the internal consistency of the brain signal \cite{schrimpf_brain-score_2018}.

To ensure that synthetic-video scoring is not confounded by generation quality, particularly resolution, we standardize all videos before final activation prediction. Each generated video is exported with FFmpeg at \(256 \times 256\) resolution, 24 fps, 2 s duration, CRF 18, medium preset, and \texttt{yuv420p} pixel format, then reloaded as tensors for encoding-model prediction. Image inputs are converted into 2-second static videos using the same format.

\section{Details of the \emph{NEvo} search procedure.}
\label{suppl:evo_pipeline}

\subsection{Hyperparameter summary}
\label{suppl:hyperparameter}

We summarize the hyperparameters used in Sec.~\ref{method:nevo} and Sec.~\ref{method:two-stage}. Unless otherwise stated, we use a population size of \(N=20\), an elite fraction of \(P_e=0.3\), a single-point crossover rate of \(P_c=0.5\), and a per-attribute mutation rate of \(P_m=0.2\). These correspond directly to the inputs of Alg.~\ref{alg:nevo}.

The population size \(N\) controls the number of candidate prompts evaluated per generation. The elite fraction \(P_e\) determines the proportion of top-scoring candidates retained as parents, enforcing selection pressure toward high-activation stimuli. The crossover rate \(P_c\) specifies the probability of recombining two parent prompts via single-point crossover, exchanging subsets of attributes to form offspring. The mutation rate \(P_m\) is applied independently to each attribute, randomly resampling it from its domain, enabling exploration of the prompt space.

For the two-stage search, we use \(T_{\mathrm{img}}=400\) evaluations for the image stage and \(T_{\mathrm{vid}}=200\) evaluations for the video stage. 
The image stage optimizes static content attributes to identify promising configurations, which initialize the video stage where motion and temporal attributes are further refined. This decomposition exploits the approximate factorization of content and dynamics, reducing the number of expensive video evaluations.
We chose these evaluation budgets because performance generally plateaued beyond these points.

\begin{figure}
    \centering
    \includegraphics[width=\linewidth]{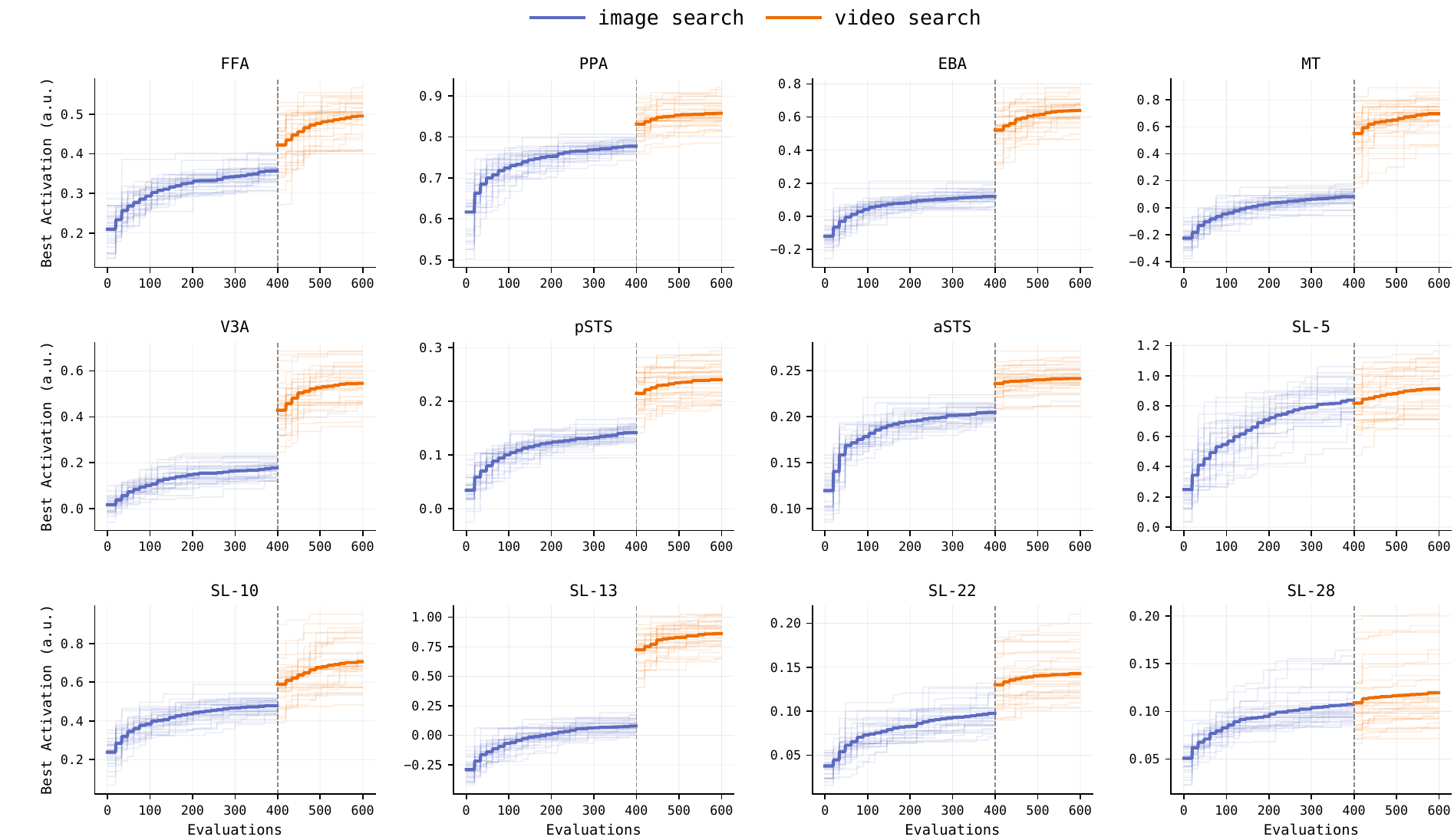}
    \caption{Two-stage search across iterations for additional ROIs.}
    \label{fig:nevo-curves}
\end{figure}

\begin{figure}
    \centering
    \includegraphics[width=0.7\linewidth]{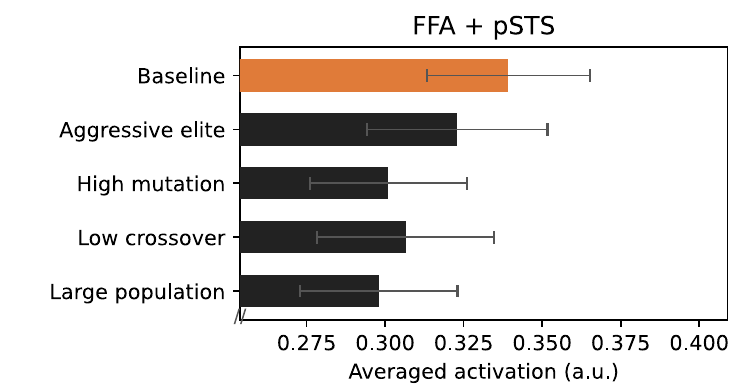}
    \caption{Hyperparameter ablation for the genetic search algorithm. Mean activation (a.u.) across FFA and pSTS, averaged over image and video search stages and 5 random seeds. Error bars show SEM. We compare five configurations: baseline (population size 20, mutation rate 0.2, crossover rate 0.5, elite fraction 0.3), high mutation (mutation rate 0.5), low crossover (crossover rate 0.2), large population (population size 50, elite fraction 0.2), and aggressive elite (mutation rate 0.1, crossover rate 0.6, elite fraction 0.5). Differences between configurations are not statistically significant across 5 seeds; we use the baseline throughout.}
    \label{fig:nevo-param-ablation}
\end{figure}

Hyperparameter sensitivity is evaluated in Fig.~\ref{fig:nevo-param-ablation}, where we vary mutation rate, crossover rate, elite fraction, and population size while keeping the remaining parameters fixed at their default values. 
Additional search results for more ROIs are shown in Fig.~\ref{fig:nevo-curves}.

\subsection{Design of the search space}
\label{suppl:search-space}

The prompt search space \(\mathcal{P}=\mathcal{P}_{\mathrm{img}}\times\mathcal{P}_{\mathrm{vid}}\) is organized into categories, where each category represents one dimension of the visual or temporal content of a stimulus. 
Options within each category span the relevant range of values for that dimension.

For the image space \(\mathcal{P}_{\mathrm{img}}\), categories are grouped by the level of the visual hierarchy they target. Low-level categories such as SpatialFrequency, EdgeDensity, and Pattern cover basic image statistics relevant for early visual cortex. Mid-level categories such as Texture and Surface cover material properties. Higher-level categories cover object shape (ObjectForm), body configuration (BodyContent, BodyAction), face identity and gaze (FaceContent, FaceViewpoint, FaceOcclusion), and social interaction (SocialDynamics, SocialCue, IntentionSignal). Scene-level categories such as SceneCategory and SceneGeometry target scene-selective regions.

The image space also includes motion-related categories such as ScenePhysics and MotionSourceType. Static images cannot depict motion directly, but they can imply it through cues like frozen mid-air poses or dynamic scene composition. These categories allow the image search to find content that implies motion, serving as a useful initialization for regions that care about dynamic features. The video search then builds on this by animating the anchor image with explicit temporal dynamics.

Not all categories are directly motivated by known functional selectivity. Categories like Mood, Style, and RenderType serve as quality anchors that help the generative model produce coherent images, and as exploratory dimensions that give the search freedom to discover effective combinations beyond our prior expectations.

The video motion space \(\mathcal{P}_{\mathrm{vid}}\) covers temporal dimensions only, since the anchor image already provides visual content in I2V mode. Categories cover motion source and scale (MotionSourceType, MotionScale), speed and rhythm (MotionStrength, TemporalRhythm), the primary action (PrimaryAction), and how agents interact over time (Interaction, SocialContingency, EventStructure).

The image and video motion search spaces are summarized in Tables~\ref{tab_image_space_summary} and~\ref{tab_video_space_summary}, with the complete listings provided in Tables~\ref{tab_image_space_full} and~\ref{tab_video_space_full}. 

In Figs.~\ref{fig:f3}--\ref{fig:f4}, we constrain the search space to categories relevant to each target ROI. Categories broadly aligned with the known selectivity of the ROI are kept active, while all other categories are fixed to their empty option. This reduces the effective search dimensionality and focuses the search budget on informative attributes.
For the lateral-stream analysis in Fig.~\ref{fig:f5}, we use the full unconstrained search space at all searchlight locations, allowing \emph{NEvo} to discover effective feature combinations without imposing prior assumptions about regional preferences.

\section{Automated visual-content annotation for synthesis analysis}
\label{suppl:property-annotation}

To characterize the visual content of both the synthesized and naturalistic videos, 
we conducted an automated annotation analysis using a video understanding model 
(Gemini-2.5-Flash). For each video, the model was prompted to return a brief 
one-sentence summary of the visible content and to rate a set of perceptual and 
social properties on a standardized numerical scale.

\paragraph{Annotation prompt.}
The model was given the following structured instruction:

\begin{quote}
\textit{You are a video annotation assistant. Your task is to rate a short video 
clip on a set of perceptual and social properties. Return strict JSON only — no 
markdown fences, no prose outside the JSON object. The JSON must contain exactly 
two keys: \texttt{summary} (a $\leq$35-word description of the visible content) 
and \texttt{properties} (a JSON object with each property name as a key, each 
value containing a \texttt{score} and a $\leq$20-word \texttt{evidence} string 
grounded in specific visible cues). All properties use a 0–10 integer scale 
(0 = absent/minimal, 10 = extreme/maximal), except \texttt{person\_count\_estimate} 
which uses values 0–5 only. Social properties are set to 0 when fewer than two 
agents are visible. Each score should be based solely on visible content in the 
clip, rating the full clip rather than a single frame.}
\end{quote}

\paragraph{Annotation properties.}
The full set of annotated visual properties is listed in 
Table~\ref{tab_annotation_properties}, organized by functional category. 
Properties span low-level visual features (e.g., spatial complexity, texture), 
scene-level content (e.g., spatial expanse, navigation), motion (e.g., biological 
motion, camera motion), action (e.g., goal-directed action), human/body content 
(e.g., face and body presence), and social interaction (e.g., joint action, 
communication).

\begin{figure*}
    \centering
    \includegraphics[width=\linewidth]{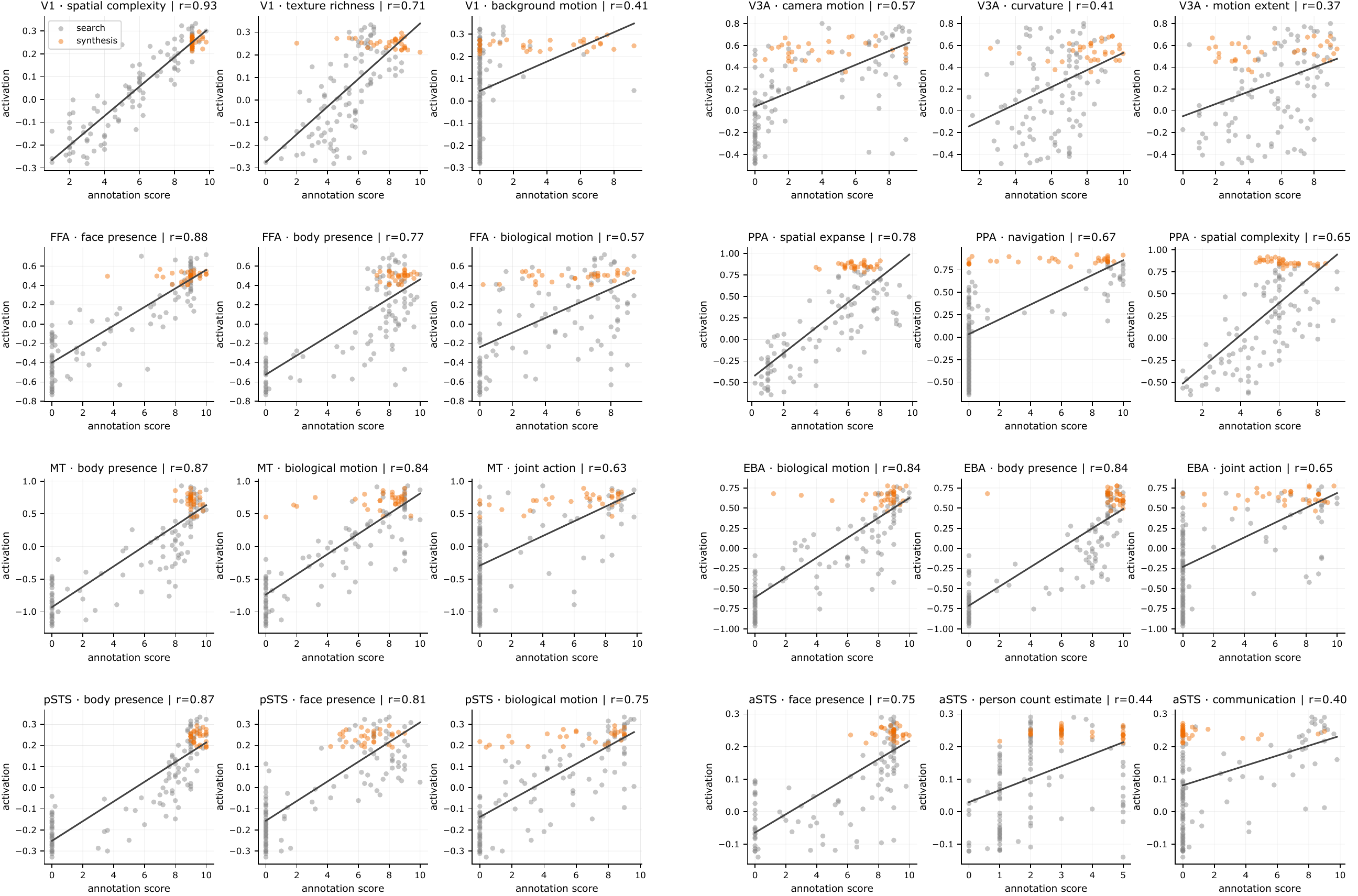}
    \caption{Correlation between model-predicted activation and annotated visual 
    properties across all ROIs. For each ROI, the three visual properties with the 
    highest Pearson correlation are shown. Gray points represent naturalistic videos 
    sampled uniformly across the full range of predicted activation; orange points 
    represent synthesis stimuli.}
    \label{fig:searchlight-more-correlations}
\end{figure*}

\paragraph{Visual property--activation correlation.}
For each target ROI, we sampled a set of naturalistic videos ($n = 100$) drawn 
uniformly across the full range of model-predicted activation (from lowest to 
highest) from the Moments in Time \cite{monfort_moments_2019} training set (after excluding 11{,}971 videos with no visual content, e.g., entirely black frames; $n = 715{,}334$ usable videos), and synthesized a matched set of videos ($n = 32$) using NEvo. 
This procedure was applied to eight ROIs (V1, V3A, FFA, PPA, MT, EBA, pSTS, 
and aSTS), yielding a total of 1{,}052 test videos. 
All videos were annotated using the prompt described above. 
For each ROI, we computed the Pearson correlation between each annotated visual 
property score and the corresponding model-predicted activation, and report 
results for the representative properties (Fig.~\ref{fig:f5}B) and top-correlated properties (Fig.~\ref{fig:searchlight-more-correlations}) per ROI. Full results 
across all properties and ROIs are provided in Table~\ref{tab_property_activation_correlation}.

\paragraph{Reliability of annotation.}
To assess the stability of Gemini-2.5-Flash annotations, each of the 1{,}052 
test videos was annotated five times using the same prompt, with a temperature 
of 0.2 to promote deterministic outputs. Test-retest reliability was evaluated 
using the intraclass correlation coefficient ICC(C,1) (single-measure 
consistency formulation\cite{mcgraw_forming_1996, koo_guideline_2016}), 
quantifying annotation consistency across five repeated runs on the same videos. Results are 
reported in Table~\ref{tab_annotation_reliability}. Final annotation scores 
were averaged across the five repeated runs before computing 
property--activation correlations.

\paragraph{Word cloud generation.}
For each searchlight cluster, we annotated $n = 32$ synthesized videos using the 
same Gemini prompt to obtain one-sentence descriptions. 
Descriptions were tokenized, and standard English stopwords as well as 
domain-specific terms related to video generation style and generic scene 
descriptors were removed before computing word frequencies.

\section{ROI construction}
\label{suppl:roi-construction}

\paragraph{Functional ROIs.}
ROIs for EBA, FFA, PPA, MT, pSTS, and aSTS were defined in group-level 
\textit{fsaverage5} surface space using individual subject localizer data 
from the social interaction dataset \cite{mcmahon_hierarchical_2023}. 
For each subject, binary ROI masks in T1w volumetric space were projected 
onto the \textit{fsnative} cortical surface using \texttt{mri\_vol2surf} 
with nearest-neighbor interpolation, and subsequently resampled to 
\textit{fsaverage5} using \texttt{mri\_surf2surf}. Both steps were performed 
using FreeSurfer 8.1.0 \cite{fischl_freesurfer_2012b} separately for each 
hemisphere. Group-level masks were then constructed by majority voting: 
a vertex was included if at least one subject had it labeled as part of 
that ROI, maximizing spatial coverage while retaining functionally relevant vertices.
Unless otherwise specified, ROI activation is computed across both hemispheres.

\paragraph{Early visual cortex ROIs.}
V1 and V3A are not covered by the functional localizers in the social interaction dataset. 
We therefore defined them using the corresponding labels from the multi-modal parcellation 
atlas of \citet{glasser_multi-modal_2016}, resampled to the \textit{fsaverage5} surface.

\begin{figure*}
    \centering
    \includegraphics[width=\linewidth]{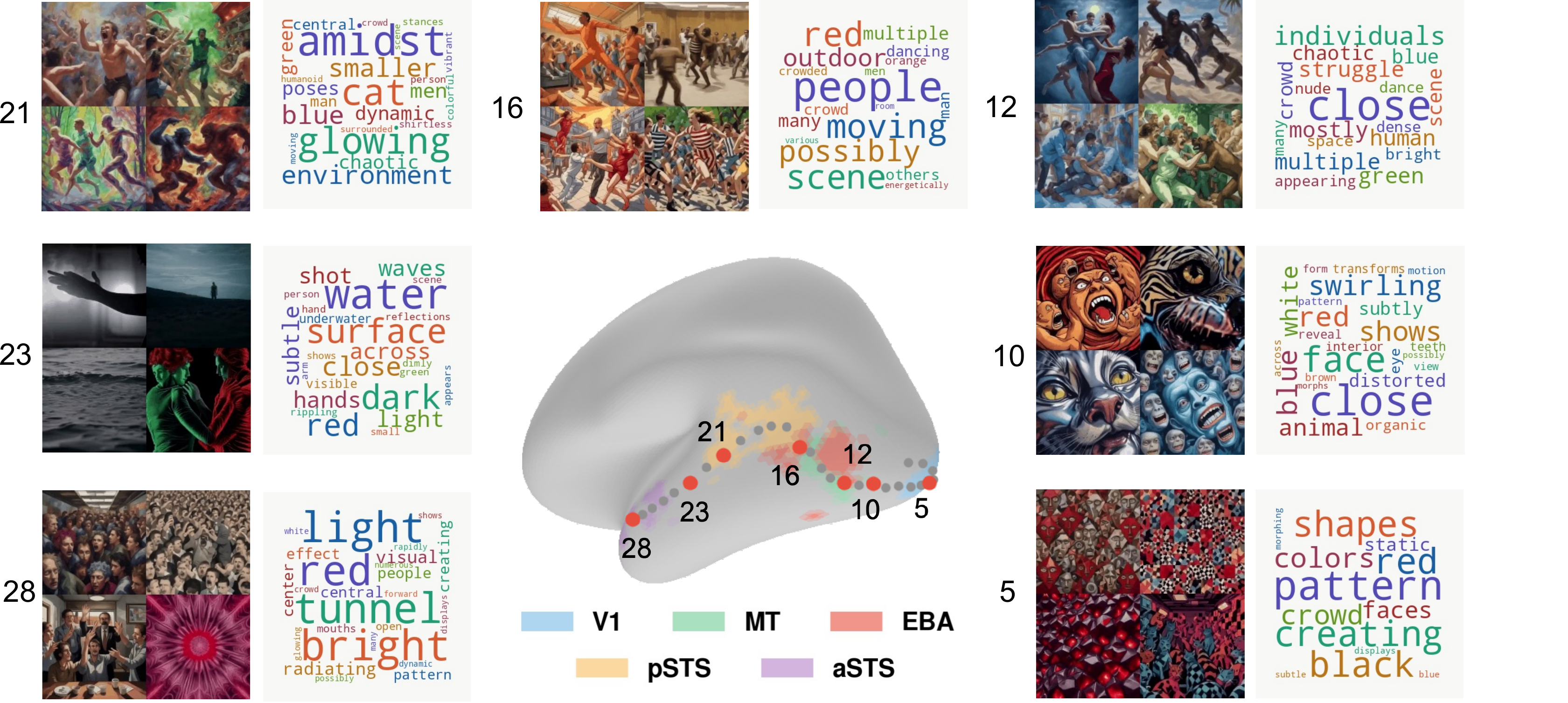}
    \caption{
    \textbf{Searchlight synthesis along the lateral visual stream in the left hemisphere.}
    Top synthesis stimuli (left) and word clouds from Gemini-generated descriptions (right) 
    are shown for example searchlight clusters along the left-hemisphere lateral stream 
    trajectory. Cluster indices are marked on the cortical surface. Feature selectivity 
    follows a similar progression to the right hemisphere, transitioning from repeated 
    colorful patterns in early visual cortex (clusters 5 and 10) to motion and body content 
    in mid-level regions (clusters 12 and 16), and physical interaction content in 
    mid-to-high-level regions (cluster 21). However, anterior STS clusters (clusters 23 
    and 28) show weaker social interaction selectivity compared to the right hemisphere, 
    potentially reflecting hemispheric lateralization: the left STS is more strongly 
    associated with language and auditory processing \cite{friederici_brain_2011a}, which are 
    absent from our model's training data, and these regions also exhibit lower noise 
    ceilings in our dataset.
    }
    \label{fig:searchlight-lh}
\end{figure*}

\paragraph{Searchlight patches.}
To characterize functional selectivity along the lateral visual stream beyond predefined ROIs, 
we constructed a sequence of searchlight patches tracing a continuous trajectory through 
landmark regions on the \textit{fsaverage5} cortical surface. 
For each hemisphere, we specified an ordered list of landmark ROIs (V1, MT, EBA, pSTS, and aSTS) 
and connected consecutive ROIs along surface geodesics, yielding a one-dimensional path 
through the lateral stream. 
Along this path, we placed searchlight centers \cite{kriegeskorte_informationbased_2006} 
at a fixed stride of $5$\,mm, and defined each patch as the set of cortical vertices within 
a $10$\,mm geodesic radius of its center.
This procedure yields 28 patches in the left hemisphere and 30 patches in the right 
hemisphere. Patches are indexed sequentially from posterior to anterior within each 
hemisphere; indices reported in Fig.~\ref{fig:f5}A and Fig.~\ref{fig:searchlight-lh} refer 
to this ordering.

\section{Controlled synthesis in non-naturalistic scenario}
\label{suppl:controlled_synthesis}

\paragraph{Synthesis methods.} 
For the controlled synthesis, we use Gemini \cite{google_gemini_2026} to generate initial images depicting two abstract objects with matched backgrounds and color schemes, and then apply the second-stage image-conditioned video search to animate them.
The search pipeline follows Suppl.~\ref{suppl:evo_pipeline} with two modifications. First, we use HunyuanVideo \cite{kong2024hunyuanvideo} as the generation model. Second, we add prompt constraints to ensure that the generated videos contain only the two objects, without extra objects or deformation into human-like figures.

\begin{figure*}
    \centering
    \includegraphics[width=\linewidth]{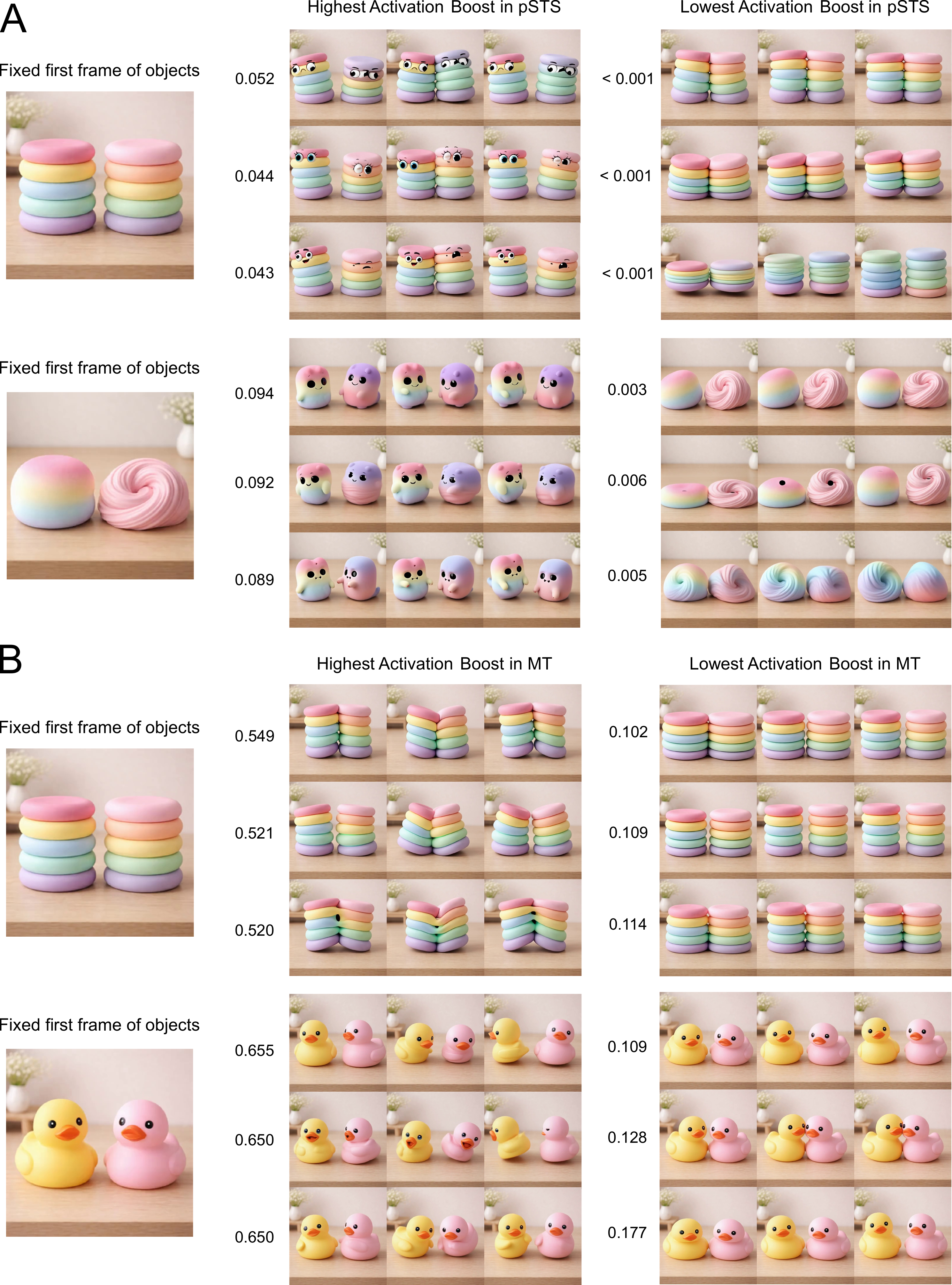}
    \caption{
    \textbf{Additional controlled-synthesis examples for pSTS and MT across multiple non-naturalistic anchors.}
    Extended results for the controlled synthesis experiment in Fig.~\ref{fig:f6}, showing different objects: plasticine stacks, deformable blobs, and rubber ducks. For each anchor, we run for multiple seeds and show the three videos with the highest predicted activation boosts (middle columns) and the three with the lowest boosts (right columns). Values indicate predicted activation improvements (a.u.) over the static first-frame baseline.
    \textbf{A. pSTS optimization.} High-boost videos consistently introduce face-like features and coordinated, interactive movements between objects, regardless of the underlying anchor, whereas low-boost videos retain the static appearance with minimal dynamics or social structure.
    \textbf{B. MT optimization.} High-boost videos exhibit motion-rich dynamics such as translation, oscillation, and deformation of the objects, while low-boost videos show little coherent motion.
    }
    \label{fig:s12}
\end{figure*}

\paragraph{Controlled synthesis results.}
Fig.~\ref{fig:s12} shows controlled-synthesis examples for pSTS and MT across multiple non-naturalistic image anchors, including plasticine stacks, deformable blobs, and rubber ducks. For each anchor, we run multiple seeds ($n = 384$) and display videos from the top and bottom 3\% of predicted activation boost relative to the static anchor. Despite prompt constraints, pSTS strongly favors face-like and interactive content, whereas MT shows weaker sensitivity to such content and is driven more by strong motion.

\section{Reproducibility}
\label{suppl:reproduce}

For each two-stage \emph{NEvo} run targeting a given ROI, we fix the random seed for both NumPy and PyTorch. Because text- and image-conditioned generation can remain partially stochastic even under fixed seeds, exact stimulus-level reproducibility is not always guaranteed: repeated runs may produce semantically similar but visually non-identical images or videos. Nevertheless, final ROI scores are expected to be stable across runs.

For each ROI, we run 32 independent seeds and aggregate the highest-scoring stimuli for analysis. A single seed is run on one A100 GPU with 80 GB memory, 16 CPU cores, and at most 64 GB CPU memory. Under the default hyperparameters in Suppl.~\ref{suppl:hyperparameter}, one full two-stage run takes approximately 6000 s.

Our supplementary files include the prompt configurations, random seeds, and algorithm code used for all reported analyses.


\newpage

\begin{table}[h]
\centering
\caption{Summary of the image prompt search space (614 options, 30 categories). Each category contributes one discrete gene to the prompt. Example options are representative samples from each category.}
\label{tab_image_space_summary}
\small
\begin{tabular}{llrp{7cm}}
\toprule
\textbf{Group} & \textbf{Category} & \textbf{\# Options} & \textbf{Example options} \\
\midrule
\multirow{4}{*}{Rendering} 
    & RenderType      & 25 & photorealistic, cinematic still, 35mm film photo, infrared photography \\
    & Lighting        & 25 & dramatic chiaroscuro, golden hour warm, neon light, cold moonlight \\
    & Style           & 24 & surrealist dreamscape, hyperrealism, gothic dark, cyberpunk noir \\
    & Color           & 33 & vivid saturated, dominant red, teal and orange, deep black shadows \\
\midrule
\multirow{4}{*}{Low-level} 
    & Pattern         & 25 & high contrast checkerboard, radial starburst, fractal branching \\
    & SpatialFrequency & 17 & high frequency texture, horizontal sine grating, random phase pattern \\
    & EdgeDensity     & 10 & dense sharp edges, sparse isolated edges, soft blurry edges \\
    & Surface         & 25 & wet glistening skin, cold brushed metal, translucent membrane \\
\midrule
\multirow{5}{*}{Face \& Body} 
    & FaceContent     & 28 & direct eye contact, baby face, angry face, multiple faces \\
    & FaceViewpoint   & 11 & direct eye contact, looking to side, unfocused distant gaze \\
    & FaceOcclusion   & 13 & full face visible, face behind mask, face in shadow \\
    & BodyContent     & 17 & full body front, hands close-up, headless body, body in fog \\
    & BodyAction      & 25 & standing still, reaching toward viewer, dancing, stumbling forward \\
    & BodyPartFocus & 10 & full body, hands and arms, legs and gait, torso posture, back and spine \\
\midrule
\multirow{5}{*}{Social} 
    & AgentType       & 23 & adult person, robot, moving triangle, silhouette person \\
    & SocialCue       & 20 & eye contact, pointing gesture, beckoning gesture, looming dominant \\
    & SocialDynamics  & 21 & one chasing other, working together, ignoring each other \\
    & IntentionSignal & 12 & reaching for goal, helping another person, hiding something \\
    & AgentNumber     &  5 & one person, two people, three people, crowd of people \\
\midrule
\multirow{4}{*}{Scene} 
    & SceneCategory   & 29 & city street, hospital room, abandoned building, snowy landscape \\
    & SceneGeometry   & 18 & deep perspective, narrow corridor, bird eye view, mirrored room \\
    & SceneOccupancy  &  7 & completely empty, densely packed, single object only \\
    & ScenePhysics    & 20 & frozen mid-air, swirling fluid vortex, gravity defying float \\
\midrule
\multirow{3}{*}{Object} 
    & ObjectForm      & 21 & everyday object, floating object, tangled objects, melting object \\
    & Texture         & 22 & rough coarse surface, reptile scales, cracked ice, honeycomb cells \\
    & HandObjectInteraction & 29 & hand throwing object, hand painting surface, hand kneading dough \\
\midrule
\multirow{3}{*}{Other} 
    & Subject         & 29 & embracing couple, figure in doorway, crowd from above \\
    & Framing         & 20 & extreme close-up, fisheye distortion, symmetrical framing \\
    & Mood            & 25 & uncanny dread, primal fear, tender warmth, violent rage \\
    & MotionSourceType & 25 & spinning wheel, rotating gear, flowing water, falling leaves\\
\midrule
\multicolumn{2}{l}{\textbf{Total}} & \textbf{614} & 30 categories\\
\bottomrule
\end{tabular}
\end{table}

\begin{table}[h]
\centering
\caption{Summary of the video motion prompt search space (163 options, 11 categories). Used in the I2V stage where an anchor image provides visual content and the search optimizes motion and interaction dynamics.}
\label{tab_video_space_summary}
\small
\begin{tabular}{lrp{9cm}}
\toprule
\textbf{Category} & \textbf{\# Options} & \textbf{Example options} \\
\midrule
MotionSourceType    & 25 & person walking, animal moving, bouncing ball, exploding particles \\
MotionStrength      &  9 & barely moving, moderate steady motion, very fast intense, building up speed \\
MotionScale         &  7 & single body part, whole scene moving, background only moves \\
PrimaryAction       & 25 & walking, dancing, oscillating, merging together, collapsing inward \\
Interaction         & 20 & chasing other, working together, passing object, blocking the path \\
SocialContingency   &  8 & no contingency, back and forth, moving in sync, call and response \\
EventStructure      & 12 & chase sequence, cause and effect, buildup and release, parallel events \\
CameraMotion        & 11 & static camera, tracking shot, zoom in, crane shot rising \\
TemporalRhythm      & 10 & smooth continuous, sudden bursts, long pause then move \\
AgentNaturalness    & 12 & realistic person, geometric shape, point light dots, shadow only \\
HandObjectInteraction & 24 & hand throwing and releasing, hand spinning object, hand kneading repeatedly \\
\midrule
\textbf{Total}      & \textbf{163} & 11 categories \\
\bottomrule
\end{tabular}
\end{table}

\clearpage
\begin{center}
\begin{longtable}{lp{11cm}}
\caption{Full image prompt search space.}
\label{tab_image_space_full} \\
\toprule
\textbf{Category} & \textbf{Options} \\
\midrule
\endhead
\bottomrule
\endfoot

RenderType &
photorealistic,
cinematic still,
national geographic photo,
extreme macro photo,
drone top-down view,
35mm film photo,
digital concept art,
dark fantasy art,
unreal engine render,
double exposure photo,
fashion editorial photo,
infrared photography,
underwater photography,
medical illustration,
security camera footage,
vintage film photo,
polaroid snapshot,
long exposure photo,
tilt shift photo,
scanner photography,
pinhole camera photo,
thermal imaging photo,
x-ray photograph,
electron microscope image,
night vision footage \\
\midrule
 
Subject &
portrait face,
direct eye contact,
smiling face,
fearful wide eyes,
angry bared teeth,
embracing couple,
mother and newborn,
hostile crowd,
two figures fighting,
dancer mid-motion,
sprinting athlete,
hands reaching out,
falling figure,
body silhouette,
predator close-up,
snake striking,
long dark corridor,
spiral staircase,
crashing wave,
volcanic eruption,
lone figure walking,
crowd from above,
child playing,
elderly person sitting,
figure in doorway,
person in window,
hands clasped together,
figure underwater,
person in fog \\
\midrule
 
Lighting &
dramatic chiaroscuro,
golden hour warm,
harsh midday sun,
rim light edge,
backlit silhouette,
volumetric god rays,
neon light,
strobe freeze frame,
softbox studio,
single spotlight,
cold moonlight,
eerie underlighting,
natural window light,
dappled forest light,
emergency red light,
candlelight warm flicker,
fluorescent office light,
overcast flat light,
firelight orange glow,
dawn pink light,
deep shadow low key,
high key bright white,
colored gel light,
light through blinds,
underwater caustic light \\
\midrule
 
Mood &
overwhelming awe,
uncanny dread,
eerie unease,
euphoric joy,
primal fear,
serene calm,
epic grandeur,
deep melancholy,
ominous threat,
suspenseful tension,
disgust revulsion,
urgent danger,
hypnotic trance,
body horror,
sensory overload,
quiet loneliness,
tender warmth,
playful energy,
cold detachment,
obsessive fixation,
nostalgic wistfulness,
claustrophobic panic,
dreamlike confusion,
violent rage,
peaceful serenity \\
\midrule
 
Color &
vivid saturated,
muted desaturated,
monochromatic,
sepia warm,
teal and orange,
complementary contrast,
deep black shadows,
neon electric,
earthy brown tones,
soft pastel,
rich jewel tones,
warm orange tones,
cool purple tones,
high contrast split,
faded washed out,
overexposed bright,
underexposed dark,
duotone two colors,
acid green neon,
deep navy blue,
black and white,
red and green,
blue and yellow,
dominant red,
dominant blue,
dominant green,
dominant yellow,
dominant orange,
dominant purple,
pink and cyan,
gold and black,
red and black,
white and silver \\
\midrule
 
Style &
surrealist dreamscape,
high realism,
cyberpunk noir,
vaporwave aesthetic,
Bauhaus geometric,
baroque theatrical,
psychedelic,
hyperrealism,
pop art,
street art graffiti,
impressionist painterly,
minimalist clean,
gothic dark,
futurist dynamic,
brutalist concrete,
art nouveau organic,
expressionist distorted,
flat design,
glitch art,
ukiyo-e woodblock,
retro sci-fi,
industrial gritty,
ethereal soft,
noir black white \\
\midrule
 
Framing &
centered subject,
extreme close-up,
wide establishing shot,
dutch angle tilt,
vanishing point,
bird's eye view,
worm's eye view,
first-person view,
over the shoulder,
reflection in water,
fisheye distortion,
rule of thirds,
subject at edge,
two-shot framing,
silhouette framing,
foreground blur,
through frame shot,
symmetrical framing,
deep depth of field,
shallow depth of field \\
\midrule
 
Pattern &
tessellated hexagons,
recursive mandala,
circuit board traces,
concentric ripple rings,
fractal branching,
sine wave interference,
geometric grid,
high contrast checkerboard,
vertical bar grating,
horizontal bar grating,
diagonal stripes,
random dot noise,
op art illusion,
spiral vortex pattern,
radial starburst,
honeycomb cells,
Voronoi cell pattern,
tangled line field,
moiré interference,
Truchet tile pattern,
reaction diffusion spots,
dendritic branching,
web or net pattern,
brick wall pattern,
fish scale pattern \\
\midrule
 
Surface &
wet glistening skin,
cold brushed metal,
rough weathered stone,
soft velvet fabric,
cracked dry earth,
sheer silk fabric,
translucent membrane,
heavy film grain,
liquid chrome mercury,
flowing molten lava,
wispy smoke,
subsurface skin scatter,
polished white bone,
rusted corroded metal,
frosted ice surface,
rough tree bark,
smooth river pebble,
worn leather texture,
bubble foam surface,
sandy desert texture,
moss covered stone,
shattered broken glass,
peeling painted wall,
woven basket texture,
oily iridescent surface \\
\midrule
 
ScenePhysics &
completely frozen still,
slow gentle drift,
explosive sudden burst,
slow motion,
frozen mid-air,
swirling fluid vortex,
impact collision,
falling descent,
chaotic turbulence,
rhythmic pulsing,
time lapse,
gravity defying float,
rapid spinning,
elastic bouncing,
shockwave expanding,
cascading collapse,
spiral inward pull,
zigzag erratic path,
smooth gliding,
violent shaking \\
\midrule
 
FaceContent &
close-up portrait,
side profile face,
three-quarter face,
upside-down face,
eyes closed face,
direct eye contact,
looking away,
baby face,
elderly face,
dog face,
cat face,
monkey face,
cartoon face,
mannequin face,
neutral expression,
angry face,
fearful face,
smiling face,
surprised face,
crying face,
multiple faces,
no face,
face in mirror,
face in crowd,
blurred face,
painted face,
face with glasses,
face half shadow \\
\midrule
 
FaceViewpoint &
direct eye contact,
slightly looking away,
looking to side,
looking down,
looking up,
eyes closed,
wide open eyes,
squinting eyes,
gaze at object,
eyes darting,
unfocused distant gaze \\
\midrule
 
FaceOcclusion &
full face visible,
half face occluded,
face behind mask,
face in shadow,
face behind glass,
digital avatar face,
face out of focus,
face at distance,
face behind bars,
face under water,
face behind smoke,
face with sunglasses,
face behind hair \\
\midrule
 
BodyContent &
full body front,
full body side,
full body from behind,
torso and arms,
legs and feet,
hands close-up,
feet close-up,
body outline only,
headless body,
body in action,
resting body,
body in shadow,
body underwater,
body in fog,
body from above,
body curled up,
body stretched out \\
\midrule
 
BodyAction &
standing still,
reaching toward viewer,
reaching for object,
walking,
running,
jumping,
falling,
crouching,
punching or kicking,
hugging,
pointing,
waving,
throwing,
pushing,
pulling,
dancing,
crawling,
sitting,
lying down,
climbing,
spinning around,
stumbling forward,
recoiling backward,
kneeling down,
stretching out \\
\midrule
 
BodyPartFocus &
full body,
hands and arms,
legs and gait,
torso posture,
shoulders and neck,
feet and ankles,
fingers close-up,
back and spine,
hip and waist,
single hand \\
\midrule
 
AgentType &
adult person,
baby or toddler,
chimpanzee or monkey,
dog,
cat,
bird,
moving triangle,
moving circle,
moving square,
stick figure,
robot,
hand only,
masked person,
silhouette person,
spider,
insect,
horse,
deer,
fish,
shark,
snake,
bear,
wolf \\
\midrule
 
SocialCue &
eye contact,
joint attention,
pointing gesture,
reaching toward person,
turning toward person,
turning away,
mirroring posture,
open arms welcome,
crossed arms closed,
protective gesture,
beckoning gesture,
pushing away,
bowing submissive,
looming dominant,
shrugging gesture,
nodding agreement,
shaking head no,
covering face,
hands on hips,
leaning in close \\
\midrule
 
SocialDynamics &
working together,
competing against each other,
one person dominant,
threatening aggressive,
playing together,
parent and child,
one chasing other,
moving in sync,
ignoring each other,
hugging affectionate,
fighting,
teaching showing,
imitating copying,
greeting handshake,
walking away,
alone single person,
comforting each other,
negotiating bargaining,
celebrating together,
mourning together,
strangers passing \\
\midrule
 
IntentionSignal &
reaching for goal,
ambiguous unclear action,
trying and failing,
accidental slip,
helping another person,
blocking another person,
asking or requesting,
offering or giving,
hiding something,
searching for something,
warning someone,
inviting someone \\
\midrule
 
AgentNumber &
one person,
two people,
three people,
small group,
crowd of people \\
\midrule
 
SceneCategory &
living room,
outdoor landscape,
city street,
forest,
beach,
kitchen,
bedroom,
office,
hallway corridor,
rooftop,
cave,
mountain,
desert,
underwater,
parking lot,
subway station,
playground,
hospital room,
church or temple,
stadium or arena,
train or bus,
elevator interior,
stairwell,
empty warehouse,
abandoned building,
snowy landscape,
swamp or marsh,
library interior,
market or bazaar \\
\midrule
 
SceneGeometry &
deep perspective,
flat surface view,
low ceiling room,
wide open panorama,
bird eye view,
looking up view,
cluttered room,
empty sparse room,
narrow corridor,
doorway framing,
window framing,
symmetrical room,
staircase,
curved arching space,
mirrored room,
split level space,
open field horizon,
tight enclosed corner \\
\midrule
 
SceneOccupancy &
completely empty,
single object only,
sparsely populated,
moderately occupied,
densely packed,
one person only,
crowded no space \\
\midrule
 
ObjectForm &
everyday object,
abstract geometric shape,
organic natural object,
tool or instrument,
household item,
food item,
vehicle,
furniture,
clothing item,
plant or flower,
transparent glass object,
reflective shiny object,
broken damaged object,
single isolated object,
multiple objects,
oversized giant object,
tiny miniature object,
floating object,
melting object,
tangled objects,
stacked objects \\
\midrule
 
Texture &
fine detailed texture,
rough coarse surface,
smooth glossy surface,
human skin,
flat solid color,
fur,
fabric,
wood grain,
stone surface,
brushed metal,
water ripples,
sand,
tree bark,
reptile scales,
feathers,
velvet,
bubble wrap,
honeycomb cells,
tangled hair,
crumpled paper,
cracked ice,
woven fiber \\
\midrule
 
SpatialFrequency &
low frequency blobs,
mid frequency edges,
high frequency texture,
white noise field,
pink noise field,
horizontal sine grating,
vertical sine grating,
diagonal sine grating,
checkerboard pattern,
random phase pattern,
coarse grid pattern,
fine dot pattern,
blurred soft field,
sharp crisp edges,
mixed frequency noise,
radial frequency pattern,
concentric ring grating \\
\midrule
 
EdgeDensity &
sparse isolated edges,
moderate edge density,
dense sharp edges,
edge-free flat field,
soft blurry edges,
high contrast edges,
broken fragmented edges,
curved smooth edges,
jagged rough edges,
layered overlapping edges \\
\midrule
 
MotionSourceType &
spinning wheel,
rotating gear,
flowing water,
falling leaves,
rising smoke,
bouncing ball,
swinging pendulum,
flapping bird wings,
waving flag,
drifting clouds,
rolling vehicle,
exploding particles,
hand movement,
fish swimming,
rotating fan blades,
falling snow,
rolling waves,
flickering flame,
blowing grass,
spinning top,
dripping water,
crawling insect,
slithering snake,
galloping horse,
tumbling rocks \\
\midrule
 
HandObjectInteraction &
hand gripping tool,
hand pouring liquid,
hand writing or drawing,
hand typing on keyboard,
hand throwing object,
hand catching object,
hand cutting or slicing,
hand molding clay,
hand playing instrument,
hand turning knob,
hand pulling rope,
hand pushing button,
hand stacking blocks,
hand tearing paper,
hand folding fabric,
hand stirring bowl,
hand picking up object,
hand placing object down,
hand opening container,
hand closing container,
hand threading needle,
hand striking match,
hand squeezing object,
hand painting surface,
hand swiping screen,
hand flipping page,
hand kneading dough,
hand snapping fingers,
hand holding nothing \\
 
\end{longtable}
\end{center}

\begin{center}

\begin{longtable}{lp{11cm}}
\caption{Full video motion search space.}
\label{tab_video_space_full} \\
\toprule
\textbf{Category} & \textbf{Options} \\
\midrule
\endhead
\bottomrule
\endfoot
 
MotionSourceType &
person walking,
person running,
animal moving,
spinning wheel,
rotating gear,
flowing water,
falling leaves,
rising smoke,
bouncing ball,
swinging pendulum,
flapping bird wings,
waving flag,
drifting clouds,
rolling vehicle,
exploding particles,
falling snow,
rolling ocean waves,
flickering flame,
blowing grass field,
spinning top,
dripping water,
crawling insect,
galloping horse,
tumbling rocks,
rotating windmill \\
\midrule
 
MotionStrength &
barely moving,
very subtle motion,
gentle slow motion,
moderate steady motion,
strong fast motion,
very fast intense,
intermittent bursts,
building up speed,
slowing to stop \\
\midrule
 
MotionScale &
single body part,
one person moving,
two entities moving,
whole scene moving,
camera moving only,
background only moves,
foreground only moves \\
\midrule
 
PrimaryAction &
walking,
running,
jumping,
dancing,
fighting,
reaching,
grasping,
rotating,
bouncing,
sliding,
flowing,
scattering,
falling,
rising,
spinning,
crawling,
swaying,
exploding outward,
collapsing inward,
stretching,
shrinking,
splitting apart,
merging together,
flickering,
oscillating \\
\midrule
 
Interaction &
no interaction,
mutual attention,
looking at object,
copying each other,
working together,
competing against,
caring for other,
threatening other,
chasing other,
fleeing from other,
touching each other,
passing object,
approaching together,
moving apart,
circling each other,
colliding,
pulling apart,
pushing together,
following behind,
blocking the path \\
\midrule
 
SocialContingency &
no contingency,
weak contingency,
clear response,
back and forth,
taking turns,
moving in sync,
mirroring movement,
call and response \\
\midrule
 
EventStructure &
continuous process,
single event,
repeating cycle,
approach then contact,
chase sequence,
cause and effect,
escalating sequence,
two things at once,
transformation sequence,
buildup and release,
start stop start,
parallel events \\
\midrule
 
CameraMotion &
static camera,
slow pan left,
slow pan right,
tracking shot,
zoom in,
zoom out,
orbiting around,
handheld shaky,
slow tilt up,
slow tilt down,
crane shot rising \\
\midrule
 
TemporalRhythm &
smooth continuous,
sudden bursts,
rhythmic periodic,
chaotic irregular,
slow then fast,
fast then slow,
pulsing in beat,
long pause then move,
constant acceleration,
constant deceleration \\
\midrule
 
AgentNaturalness &
realistic person,
cartoon character,
geometric shape,
robot or android,
animal agent,
stick figure,
shadow only,
silhouette only,
point light dots,
puppet or marionette,
masked figure,
faceless figure \\
\midrule
 
HandObjectInteraction &
hand gripping and lifting,
hand pouring liquid,
hand throwing and releasing,
hand catching object,
hand cutting or slicing,
hand molding or shaping,
hand playing instrument,
hand turning or twisting,
hand pulling toward body,
hand pushing away,
hand stacking or building,
hand tearing apart,
hand stirring or mixing,
hand picking up,
hand placing down carefully,
hand opening or closing,
hand squeezing and releasing,
hand painting or brushing,
hand swiping across,
hand kneading repeatedly,
hand snapping fingers,
hand rolling object,
hand spinning object,
no hand interaction \\
 
\end{longtable}
\end{center}

\newpage
\begin{table}[h]
\centering
\caption{Visual properties annotated by Gemini-2.5-Flash, organized by 
functional category. All properties use a 0--10 integer scale unless 
otherwise noted.}
\label{tab_annotation_properties}
\small
\begin{tabular}{llp{8.5cm}}
\toprule
\textbf{Category} & \textbf{Property} & \textbf{Definition} \\
\midrule
\multirow{4}{*}{Low-level}
  & spatial\_complexity  
    & Degree of visual clutter or density of fine-grained patterns in the 
      spatial layout (0 = nearly empty; 10 = densely packed). \\
  & color\_richness      
    & Diversity and saturation of colors present 
      (0 = monochromatic/desaturated; 10 = wide variety of vivid colors). \\
  & texture\_richness    
    & Richness and variation of surface textures 
      (0 = smooth/plain; 10 = highly detailed and varied). \\
  & curvature            
    & Prevalence of curved versus angular shapes and contours 
      (0 = all straight/angular; 10 = predominantly rounded/curved). \\
\midrule
\multirow{2}{*}{Scene}
  & spatial\_expanse     
    & Spatial scale of the depicted environment 
      (0 = extremely confined/close-up; 10 = wide open landscape). \\
  & navigation           
    & Degree to which the video conveys first-person movement through 
      space (0 = none; 10 = strong traversal cue). \\
\midrule
\multirow{4}{*}{Motion}
  & motion\_extent       
    & Overall magnitude and spatial scale of visual motion in the frame 
      (0 = completely static; 10 = large-scale continuous motion). \\
  & biological\_motion   
    & Presence of fluid, coordinated whole-body movement 
      (0 = absent/stiff; 10 = dynamic coordinated motion). \\
  & camera\_motion       
    & Degree of camera movement during the clip 
      (0 = static; 10 = rapid large-scale camera motion). \\
  & background\_motion   
    & Motion of the scene background independent of foreground subjects 
      (0 = static; 10 = continuous large-scale background motion). \\
\midrule
\multirow{2}{*}{Action}
  & object\_directedness 
    & Extent to which a person performs sustained actions on a physical 
      object (0 = none; 10 = clear object-directed action). \\
  & goal\_directed\_action 
    & Extent to which an action is directed toward a specific functional 
      goal with a clear outcome (0 = none; 10 = clear goal-directed action). \\
\midrule
\multirow{3}{*}{Human / Body}
  & face\_presence       
    & Visibility and prominence of human faces in the frame 
      (0 = no face; 10 = face(s) occupying large portion of frame). \\
  & body\_presence       
    & Visibility and prominence of human bodies or body parts 
      (0 = none; 10 = body/bodies occupying large portion of frame). \\
  & person\_count\_estimate 
    & Number of distinct individuals present 
      (integer: 0 = none, 1–4 = exact count, 5 = five or more). \\
\midrule
\multirow{6}{*}{Social}
  & agent\_distance      
    & Physical proximity between individuals 
      (0 = very close; 10 = spatially distant; 0 if $<$2 agents). \\
  & facingness           
    & Extent to which individuals face each other 
      (0 = back-to-back; 10 = directly face-to-face; 0 if $<$2 agents). \\
  & physical\_contact    
    & Degree of physical touch between individuals 
      (0 = no contact; 10 = full-body contact; 0 if $<$2 agents). \\
  & joint\_action        
    & Extent of engagement in a joint action together 
      (0 = entirely independent; 10 = fully joint action; 0 if $<$2 agents). \\
  & synchronized\_movement 
    & Degree of temporal synchrony between individuals' movements 
      (0 = asynchronous; 10 = precisely synchronized; 0 if $<$2 agents). \\
  & communication        
    & Extent of active communication between individuals via speech, 
      gesture, or eye contact (0 = none; 10 = salient; 0 if $<$2 agents). \\
\midrule
Other
  & text\_presence       
    & Visual prominence of text or written characters in the frame 
      (0 = none; 10 = text is the dominant visual element). \\
\bottomrule
\end{tabular}
\end{table}

\begin{table}[h]
\centering
\caption{Test-retest reliability of Gemini-2.5-Flash annotations across 
five repeated runs. Reliability was evaluated using the ICC(C,1) 
single-measure consistency formulation \cite{mcgraw_forming_1996}, with 
interpretation thresholds of $<0.50$ poor, $0.50$--$0.75$ moderate, 
$0.75$--$0.90$ good, and $>0.90$ excellent \cite{koo_guideline_2016}.}
\label{tab_annotation_reliability}
\small
\begin{tabular}{llccc}
\toprule
\textbf{Category} & \textbf{Property} & \textbf{ICC(C,1)} & \textbf{95\% CI} & \textbf{Interpretation} \\
\midrule
\multirow{4}{*}{Low-level}
  & spatial\_complexity    & 0.935 & [0.929, 0.941] & Excellent \\
  & color\_richness        & 0.938 & [0.932, 0.943] & Excellent \\
  & texture\_richness      & 0.876 & [0.865, 0.887] & Good      \\
  & curvature              & 0.881 & [0.870, 0.891] & Good      \\
\midrule
\multirow{2}{*}{Scene}
  & spatial\_expanse       & 0.957 & [0.953, 0.961] & Excellent \\
  & navigation             & 0.910 & [0.902, 0.918] & Excellent \\
\midrule
\multirow{4}{*}{Motion}
  & motion\_extent         & 0.895 & [0.886, 0.904] & Good      \\
  & biological\_motion     & 0.943 & [0.938, 0.948] & Excellent \\
  & camera\_motion         & 0.887 & [0.877, 0.896] & Good      \\
  & background\_motion     & 0.767 & [0.748, 0.784] & Good      \\
\midrule
\multirow{2}{*}{Action}
  & object\_directedness   & 0.945 & [0.939, 0.949] & Excellent \\
  & goal\_directed\_action & 0.914 & [0.906, 0.921] & Excellent \\
\midrule
\multirow{3}{*}{Human / Body}
  & face\_presence         & 0.975 & [0.973, 0.977] & Excellent \\
  & body\_presence         & 0.976 & [0.974, 0.978] & Excellent \\
  & person\_count\_estimate& 0.978 & [0.975, 0.980] & Excellent \\
\midrule
\multirow{6}{*}{Social}
  & agent\_distance        & 0.892 & [0.882, 0.901] & Good      \\
  & facingness             & 0.939 & [0.933, 0.944] & Excellent \\
  & physical\_contact      & 0.938 & [0.932, 0.943] & Excellent \\
  & joint\_action          & 0.940 & [0.934, 0.945] & Excellent \\
  & synchronized\_movement & 0.913 & [0.905, 0.921] & Excellent \\
  & communication          & 0.927 & [0.920, 0.933] & Excellent \\
\midrule
Other
  & text\_presence         & 0.931 & [0.924, 0.937] & Excellent \\
\bottomrule
\end{tabular}
\end{table}
\begin{table}[h]
\centering
\caption{Pearson correlation between annotated visual property scores and 
model-predicted activation for each ROI. Values are computed over $n = 100$ 
naturalistic videos sampled uniformly across the predicted activation range, 
combined with $n = 32$ NEvo-synthesized videos per ROI.}
\label{tab_property_activation_correlation}
\small
\begin{tabular}{lrrrrrrrr}
\toprule
\textbf{Property} & \textbf{V1} & \textbf{V3A} & \textbf{FFA} & \textbf{PPA} & \textbf{MT} & \textbf{EBA} & \textbf{pSTS} & \textbf{aSTS} \\
\midrule
face\_presence        &$-$0.047 & $-$0.331 &  0.881 & $-$0.699 &  0.413 &  0.599 &  0.815 &  0.755 \\
body\_presence        & $-$0.295 & $-$0.226 &  0.770 & $-$0.703 &  0.873 &  0.837 &  0.869 &  0.312 \\
biological\_motion    & $-$0.329 & $-$0.148 &  0.570 & $-$0.490 &  0.837 &  0.843 &  0.747 & $-$0.599 \\
spatial\_complexity   &  0.930 &  0.317 & $-$0.434 &  0.651 & $-$0.129 & $-$0.431 & $-$0.545 & $-$0.225 \\
spatial\_expanse      &  0.039 & $-$0.434 & $-$0.674 &  0.780 & $-$0.217 & $-$0.411 & $-$0.286 & $-$0.261 \\
joint\_action         &  0.219 & $-$0.088 &  0.342 & $-$0.244 &  0.634 &  0.646 &  0.732 &  0.070 \\
facingness            &  0.104 & $-$0.132 &  0.438 & $-$0.231 &  0.625 &  0.640 &  0.688 &  0.336 \\
person\_count\_estimate &  0.209 & $-$0.190 &  0.420 & $-$0.049 &  0.553 &  0.579 &  0.658 &  0.438 \\
curvature             &  0.309 &  0.405 &  0.343 & $-$0.730 &  0.174 &  0.253 &  0.168 & $-$0.142 \\
texture\_richness     &  0.713 &  0.260 & $-$0.357 &  0.162 & $-$0.108 & $-$0.160 & $-$0.512 & $-$0.181 \\
synchronized\_movement &  0.279 & $-$0.056 &  0.341 & $-$0.121 &  0.498 &  0.538 &  0.590 & $-$0.196 \\
physical\_contact     &  0.117 & $-$0.052 &  0.291 & $-$0.304 &  0.584 &  0.522 &  0.486 & $-$0.008 \\
motion\_extent        & $-$0.114 &  0.372 &  0.034 &  0.210 &  0.511 &  0.570 &  0.296 & $-$0.800 \\
communication         &  0.111 & $-$0.152 &  0.446 & $-$0.233 &  0.121 &  0.269 &  0.405 &  0.401 \\
camera\_motion        & $-$0.219 &  0.573 &  0.065 &  0.599 &  0.346 &  0.183 & $-$0.254 & $-$0.469 \\
background\_motion    &  0.406 &  0.272 & $-$0.225 &  0.541 & $-$0.038 & $-$0.228 & $-$0.372 & $-$0.250 \\
object\_directedness  & $-$0.401 & $-$0.098 & $-$0.034 & $-$0.418 &  0.105 & $-$0.092 & $-$0.105 & $-$0.605 \\
goal\_directed\_action & $-$0.350 & $-$0.138 & $-$0.097 & $-$0.378 &  0.381 &  0.203 & $-$0.015 & $-$0.544 \\
agent\_distance       &  0.095 & $-$0.292 &  0.037 &  0.212 &  0.159 &  0.291 &  0.386 &  0.382 \\
navigation            &  0.041 &  0.189 & $-$0.263 &  0.669 &  0.091 & $-$0.146 & $-$0.269 & $-$0.225 \\
color\_richness       & $-$0.013 &  0.127 &  0.166 & $-$0.289 &  0.243 &  0.099 &  0.294 & $-$0.192 \\
spatial\_expanse      &  0.039 & $-$0.434 & $-$0.674 &  0.780 & $-$0.217 & $-$0.411 & $-$0.286 & $-$0.261 \\
text\_presence        & $-$0.101 & $-$0.243 & $-$0.154 & $-$0.090 & $-$0.269 & $-$0.481 & $-$0.345 &  0.025 \\
\bottomrule
\end{tabular}
\end{table}

\end{document}